\documentclass[10pt,twocolumn,letterpaper]{article}

\usepackage[pagenumbers]{cvpr} 

\usepackage{graphicx}
\usepackage{amsmath}
\usepackage{amssymb}
\usepackage{booktabs}
\usepackage{courier}
\usepackage{appendix}

\usepackage[pagebackref,breaklinks,colorlinks]{hyperref}

\usepackage[capitalize]{cleveref}
\crefname{section}{Sec.}{Secs.}
\Crefname{section}{Section}{Sections}
\Crefname{table}{Table}{Tables}
\crefname{table}{Tab.}{Tabs.}

\def\cvprPaperID{XXXX} 
\def\confName{CVPR}
\def\confYear{2023}

\begin{document}

\title{ToonAging: Face Re-Aging upon Artistic Portrait Style Transfer}



\author{Bumsoo Kim$^{1,2}$, Abdul Muqeet$^2$, Kyuchul Lee$^3$, Sanghyun Seo$^{1}$\thanks{Corresponding author.}\\
$^1$Chung-Ang University, $^2$VIVE STUDIOS, $^3$Coupang\\
{\tt\small \{bumsookim, $^*$sanghyun\}@cau.ac.kr, amuqeet@vivestudios.com,  kylee287@coupang.com}}

\maketitle

\begin{abstract}
   Face re-aging is a prominent field in computer vision and graphics, with significant applications in photorealistic domains such as movies, advertising, and live streaming. Recently, the need to apply face re-aging to non-photorealistic images, like comics, illustrations, and animations, has emerged as an extension in various entertainment sectors. However, the lack of a network that can seamlessly edit the apparent age in NPR images has limited these tasks to a naive, sequential approach. This often results in unpleasant artifacts and a loss of facial attributes due to domain discrepancies. In this paper, we introduce a novel one-stage method for face re-aging combined with portrait style transfer, executed in a single generative step. We leverage existing face re-aging and style transfer networks, both trained within the same PR domain. Our method uniquely fuses distinct latent vectors, each responsible for managing aging-related attributes and NPR appearance. By adopting an exemplar-based approach, our method offers greater flexibility compared to domain-level fine-tuning approaches, which typically require separate training or fine-tuning for each domain. This effectively addresses the limitation of requiring paired datasets for re-aging and domain-level, data-driven approaches for stylization. Our experiments show that our model can effortlessly generate re-aged images while simultaneously transferring the style of examples, maintaining both natural appearance and controllability.
\end{abstract}

\section{Introduction}
Face re-aging is a problem in computer graphics that semantically changes the apparent age according to the target age. Originally, this task was conducted at high costs by expert artists in media production fields, such as film, advertisement, and TV series. After generative models showed results that were indistinguishable from real images, there were some trials to perform re-aging using generative models. Several existing studies \cite{alaluf2021only,hsu2022agetransgan,makhmudkhujaev2021ragan,makhmudkhujaev2023raganpp,zoss2022production,muqeet2023video,li2023pluralistic,yoon2023manipulation} predominantly employ StyleGAN approaches. With the given input face image, input age, and target age, re-aging methods alter the generated images to resemble the target age. However, all these methods mainly focus on the single goal of changing a person's age in photos. However, all these methods are optimized to alter the age of real individuals in photographs. As a result, such methods often struggle when applied to NPR images

Meanwhile, artistic portraits hold a significant place in our everyday life, particularly within industries like comics, animation, posters, and advertising. These portraits, which include styles ranging from caricatures to anime, constitute essentially stylized images. Similar to re-aging processes, StyleGAN-based architectures \cite{pinkney2020resolution,huang2021unsupervised,song2021agilegan,kim2022cross,li2023parsing,jang2021stylecarigan,yang2022pastiche,wang20223d,kim2023context,zhu2023few,men2022dct,lee2023fix,khowaja2023face,yang2022vtoonify,zhang2022generalizedOD,chefer2022image, back2022webtoonme} have achieved substantial success in generating high-resolution artistic portraits. This success is largely attributable to their hierarchical style control, which also facilitates transfer learning applications. Image-to-Image translation methods \cite{kim2019u, choi2020starganv2} involve developing distinct models for each domain, which requires extensive datasets and a longer preparation process compared to methods using pre-trained models, as each domain starts from scratch. These methods are primarily oriented towards style modification with limited capability for additional attribute transformation simultaneously.

\begin{figure*}[!t]
  \centering
  
  \includegraphics[width=1\linewidth]{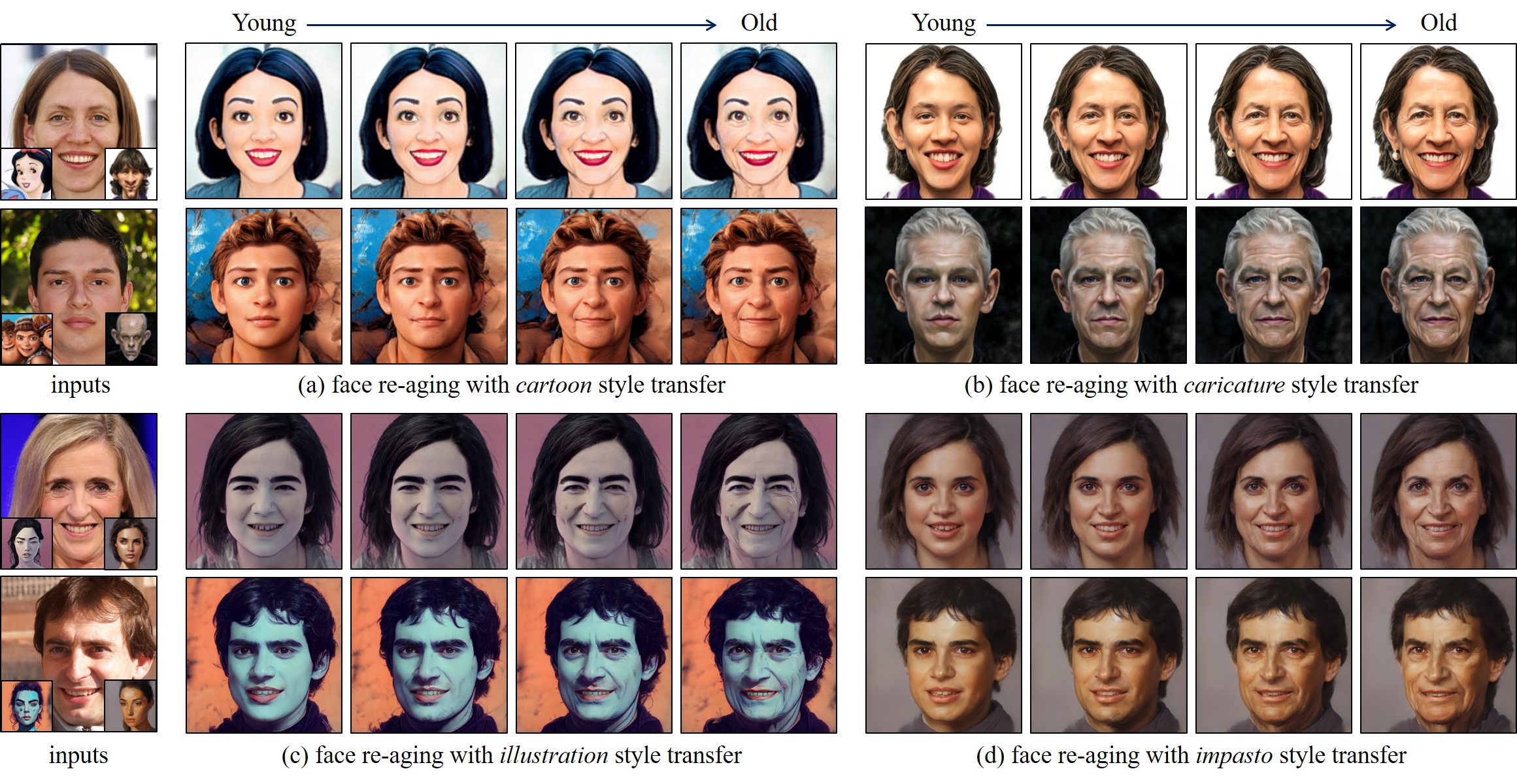}
    \caption{We propose \textit{ToonAging}, which can perform face re-aging and portrait style transfer in a single generation step. Since we adopted an exemplar-based approach, portraits can be transferred to various domains, enabling plausible re-aging progression simultaneously.}
  \label{fig:title}
\end{figure*}

\begin{figure*}[!t]
  \centering
  
  \includegraphics[width=1\linewidth]{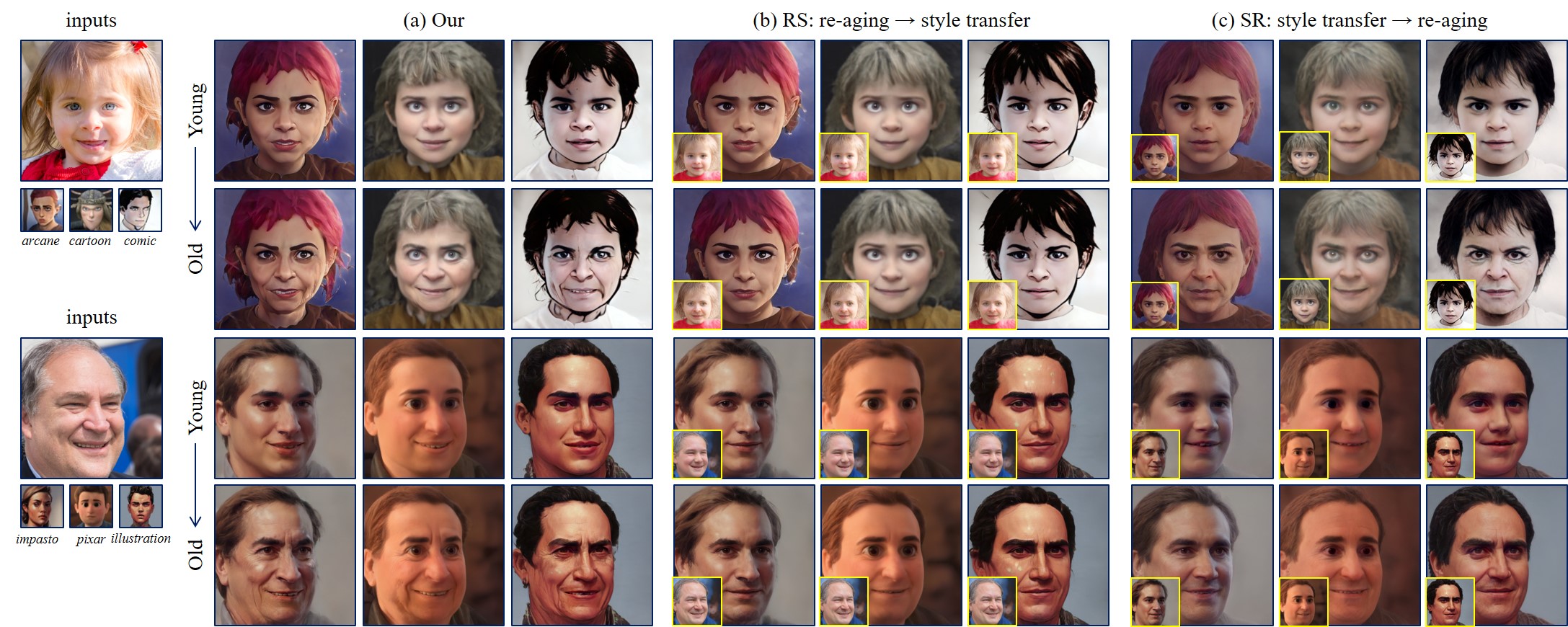}
    \caption{Comparison of our method with naive sequential approaches. (a) \textit{ToonAging} perceptually reflects the apparent age while simultaneously effectively transferring the style of $S$. (b) The naive RS (re-aging then style transfer) often loses age-related attributes, resulting in the image lacking a discernible target age and missing the facial attributes of the input (\textit{e.g.}, expression; here, the input face has a smiling expression, but the generated image does not.) (c) Conversely, SR (style transfer then re-aging) maintains a realistic style but fails to retain the input $S$'s style. Yellow boxes denote the intermediate results of the first stage in 2-stage approaches.}
  \label{fig:limitation}
\end{figure*}

As previously mentioned, despite various research efforts in both face re-aging and artistic portrait generation, to the best of our knowledge, the combination of face re-aging and artistic portraits has not been explored. The naive approach to tackle this problem is to apply each method sequentially. We investigated this approach by applying SAM \cite{alaluf2021only} to images and then applying these re-aged images with the state-of-the-art style generation technique DualStyleGAN \cite{yang2022pastiche}. The results are demonstrated in Fig. \ref{fig:limitation} (b). We encountered various issues with this approach: 1) Aging details such as wrinkles in the input image are lost. 2) DualStyleGAN does not successfully preserve facial attributes, including facial expressions like the mouth in the input image. The reason for these issues lies in the inversion processes, which do not effectively preserve the input details. We also conducted the experiment in reverse order, first applying DualStyleGAN to images and then SAM, as shown in Fig. \ref{fig:limitation} (c). We encountered the same problem of inversion with this approach as well. In this case, SAM failed to preserve the artistic details of the input images. The underlying issue is attributed to the training procedure of re-aging methods, as they are trained on \cite{or2020lifespan}, which is based on real images generated using StyleGAN \cite{karras2019stylegan}. 

To address the challenges in combining face re-aging and artistic portrait generation, we have developed a method that effectively merges latent information for each goal. Our approach combines SAM \cite{alaluf2021only} for re-aging and DualStyleGAN \cite{yang2022pastiche} for style transfer. Specifically, we extract inversions in the ${W+}$ space of a pre-trained StyleGAN network using SAM and apply these, along with a residual ${w}$, to the intrinsic path of DualStyleGAN. This combination respects the principle of superposition, allowing the effects of each network to merge seamlessly and without mutual interference. Our analysis suggests that using the ${W+}$ space, rather than the ${Z+}$ space, for the intrinsic path better preserves the subject's attributes. Additionally, by maintaining the original extrinsic path in DualStyleGAN, we naturally blend age transformation and style transfer. This unified approach offers precise control over the re-aging and style transfer processes, overcoming the limitations of previous methods and enabling the generation of more natural and realistic images. In summary, we present three primary contributions:
\begin{itemize}

\item \textit{ToonAging} innovatively combines face re-aging and portrait style transfer in a one-stage method, surpassing sequential methods in performance without needing extra datasets or training.

\item We utilize a latent fusion approach across different networks, enabling precise control over diverse aspects from coarse to fine details, including the shape, rendering style, color, and aging effects with ease.

\item In contrast to domain-level style transfer approaches, our method inherits an exemplar-based technique, allowing for style transfer across any domain without the need for additional fine-tuning or training.

\end{itemize}

\section{Related Work}
In this section, we provide a literature review covering methods for face re-aging and the generation of artistic styles in facial images.

\subsection{Face Re-Aging}

Face re-aging \cite{alaluf2021only,hsu2022agetransgan,makhmudkhujaev2021ragan,makhmudkhujaev2023raganpp,zoss2022production,muqeet2023video} including aging (i.e., \textit{Old} direction) and de-aging (i.e., \textit{Young} direction), aims to change the apparent age of a facial image to specified target age. It faces an issue where the re-aging network must perceptually preserve the original facial identity in the generated output \cite{zhai2018identity}, while simultaneously generating a significant visual effect of re-aging as well \cite{zoss2022production,muqeet2023video}. However, it is known that there is a trade-off between aging performance and identity preservation \cite{alaluf2021only,zoss2022production,muqeet2023video}. Moreover, the limited paired dataset for age labeling with corresponding faces makes the face re-aging task more difficult.

To mitigate this, \cite{zoss2022production} leverages an image-level re-aging network \cite{alaluf2021only} in a supervision manner. \cite{muqeet2023video} extends this approach to achieve superior video-level consistent re-aging output, demonstrating highly consistent re-aging effects with stable aging performance and identity preservation. Based on the StyleGAN architecture \cite{karras2019stylegan,karras2020stylegan2}, some works have explored latent vectors within learned distributions \cite{alaluf2021only,makhmudkhujaev2021ragan,makhmudkhujaev2023raganpp,gomez2022cusp,maeng2023gmba}. There have also been efforts to perform re-aging via face attribute editing \cite{tzaban2022stitch,kim2023diffusionvae,yang2023styleganex}. Recently, some researchers have attempted to address this complex training issue by employing the diffusion approach \cite{li2023pluralistic,chen2023fading}. Additionally, one study \cite{duong2019automatic} utilizes a reinforcement learning approach to identify suitable aging visuals.

\subsection{Artistic Style Face Generation}

In the field of generative models, Non-Photorealistic Rendering (NPR) style image generation \cite{kim2024minecraft, yang2022pastiche,wang2022ctlgan} and style transfer \cite{pinkney2020resolution,kim2022cross,song2023agilegan3d,shah2022multistylegan,zhou2022hrinversion,li2023multimodal,kim2023context} have emerged as appealing problems in computer graphics due to their attractive and aesthetic appearance. The first approach using generative models for face generation involves a domain-level technique called layer swapping \cite{pinkney2020resolution,kwak2022generate} which replaces source domain weights (i.e., realistic rendering) with target domain weights (i.e., NPR). Similarly, Cross-Domain Style Mixing (CDSM) \cite{kim2022cross} fuses two different domain weights at the latent level in $\mathcal{S}$ space \cite{wu2021stylespace} by fine-tuning \cite{karras2020ada} the pre-trained weights with a cartoon dataset. On the other hand, unsupervised approaches have also been employed for cartoon-style image generation \cite{huang2021unsupervised} through image-to-image translation schemes \cite{liu2017unit,zhu2017cyclegan,huang2018munit,lee2018drit,lee2020dritpp,kim2019u,xiao2022appearance,men2022unpaired,wang2022realtime}. Recently, a study leveraged a diffusion model for face stylization \cite{liu2023portrait}. StyleGAN-Fusion \cite{song2024styleganfusion} applies domain adaptation to distill the knowledge of pre-trained large-scale diffusion models, enabling the transition from the original domain to new target domains without requiring any training images in that domain. However, these approaches often require extensive dataset acquisition processes for cartoon images, leading to the problem of insufficient datasets in specific domains. To address this issue, recent work adopts an exemplar-based method \cite{yang2022pastiche} that doesn't require large datasets. Here, the key aspect of cartoon style generation is addressing the issue of large dataset requirements \cite{zoss2022production,muqeet2023video} using an exemplar-based method \cite{yang2022pastiche}, rather than a domain-level method \cite{pinkney2020resolution,kim2022cross}, which is more suitable for our task.

\begin{figure*}[!t]
  \centering
  
  \includegraphics[width=1\linewidth]{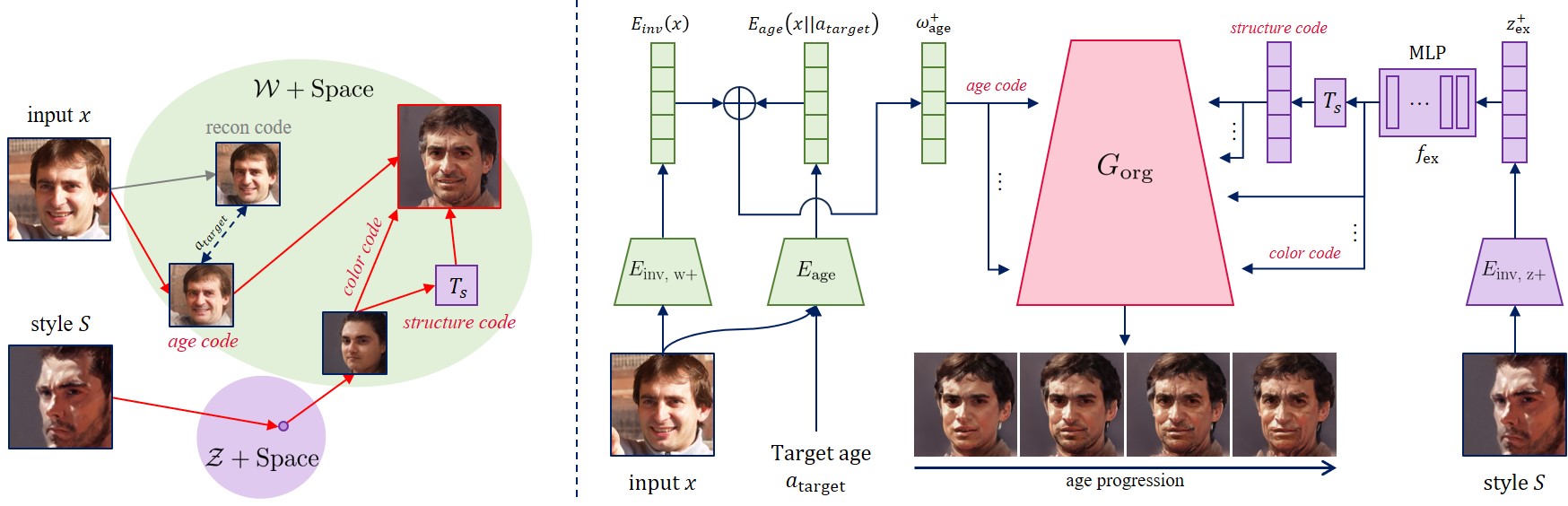}
  \caption{Our \textit{ToonAging} architecture. (Left) Schematics, (Right) Data-level architecture. Under \textit{ToonAging}, age-related attributes and age code are obtained via GAN inversion. Latent vector for reconstruction is embedded by $E_{\text{inv, w+}}$, and the age-related latent vector is embedded by $E_{\text{age}}$ as residuals to be added to the reconstruction vector. The example style $S$ is obtained by $E_{\text{inv, z+}}$ in $\mathcal{Z}+$ space, then converted into $\mathcal{W}+$ space with a learned MLP layer. Finally, the two latent vectors are used in the generator of DualStyleGAN.}
  \label{fig:toonaing_architecture}
\end{figure*}

\section{Methodology} \label{sec:method}

In the context of real-world image datasets, unsupervised learning methods for re-aging, such as those applied to datasets like FFHQ-Aging \cite{or2020lifespan}, are feasible. Additionally, data-centric approaches, as proposed in methods like \cite{zoss2022production,muqeet2023video}, can be used to create supervised learning datasets for re-aging. However, in the domain of artistic portrait images, there is a significant lack of datasets for re-aging, and the use of data-centric approaches is complex and time-consuming.

Unlike traditional methods, our approach does not rely on existing datasets or the generation of new data. Instead, we utilize existing fine-tuned networks through an exemplar-based scheme. This strategy eliminates the need for additional training by fusing individual latent vectors. For our method to be effective, it is crucial that the learned distribution and alignment within the networks are congruent. Initially, we naively applied face re-aging and portrait style transfer as a two-stage process within the StyleGAN2 latent space \cite{karras2020stylegan2} and the FFHQ distribution \cite{karras2019stylegan}. However, this approach presents its own limitations.

To perform face re-aging with portrait style transfer, we first provide a concise introduction to the existing network as preliminary knowledge in Sec. \ref{method:01}. Then, we elaborate on the details of our approach in Sec. \ref{method:02}.

\begin{figure}[t]
  \centering
  
  \includegraphics[width=1\linewidth]{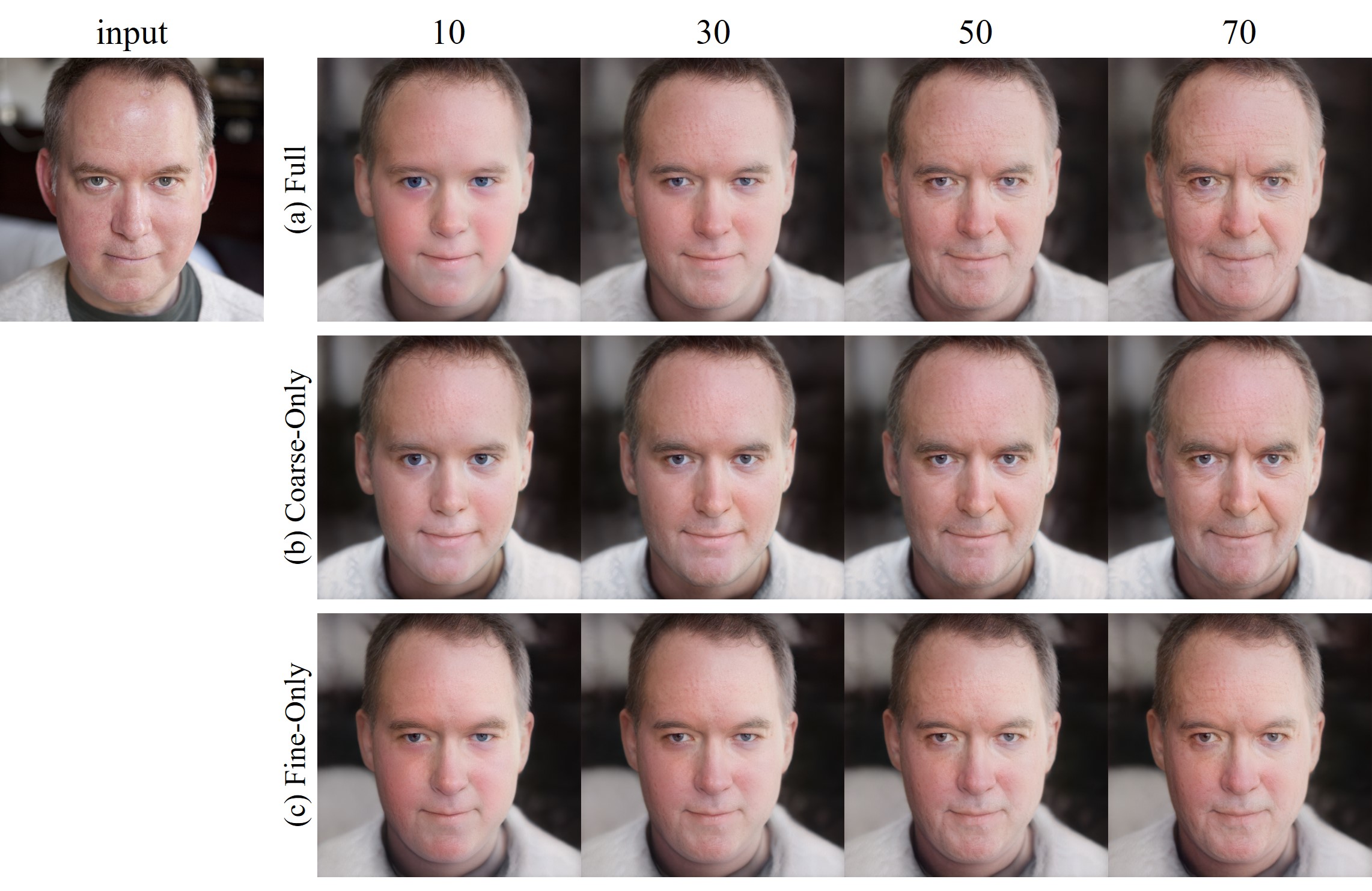}
  \caption{Latent vector behavior according to partial convolution in the generator for face re-aging. SAM was employed for latent analysis.}
  \label{fig:simulation_sam}
\end{figure}

\begin{figure}[t]
  \centering
  
  \includegraphics[width=1\linewidth]{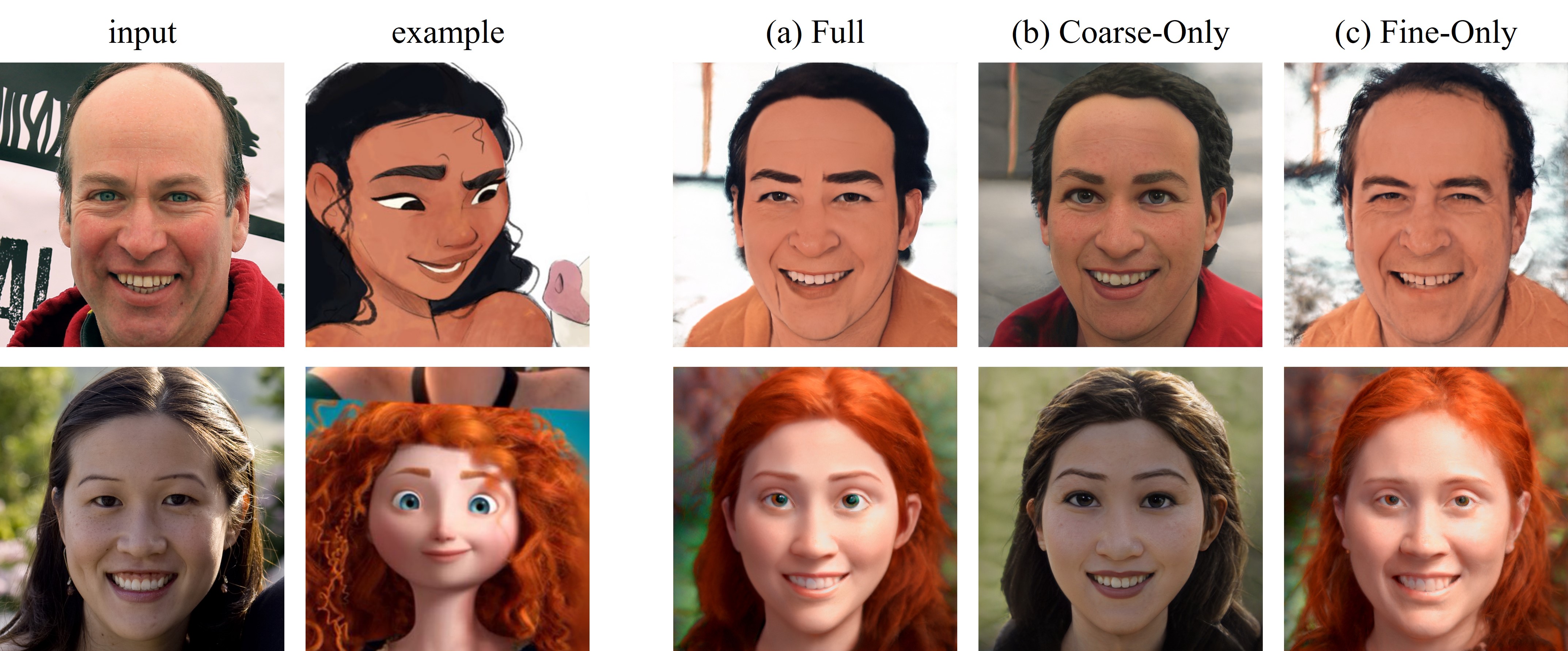}
  \caption{Latent behavior for exemplar-based style transfer. DualStyleGAN was used for latent analysis.}
  \label{fig:simulation_dualstylegan}
\end{figure}

\begin{figure*}[t]
  \includegraphics[width=0.95\linewidth]{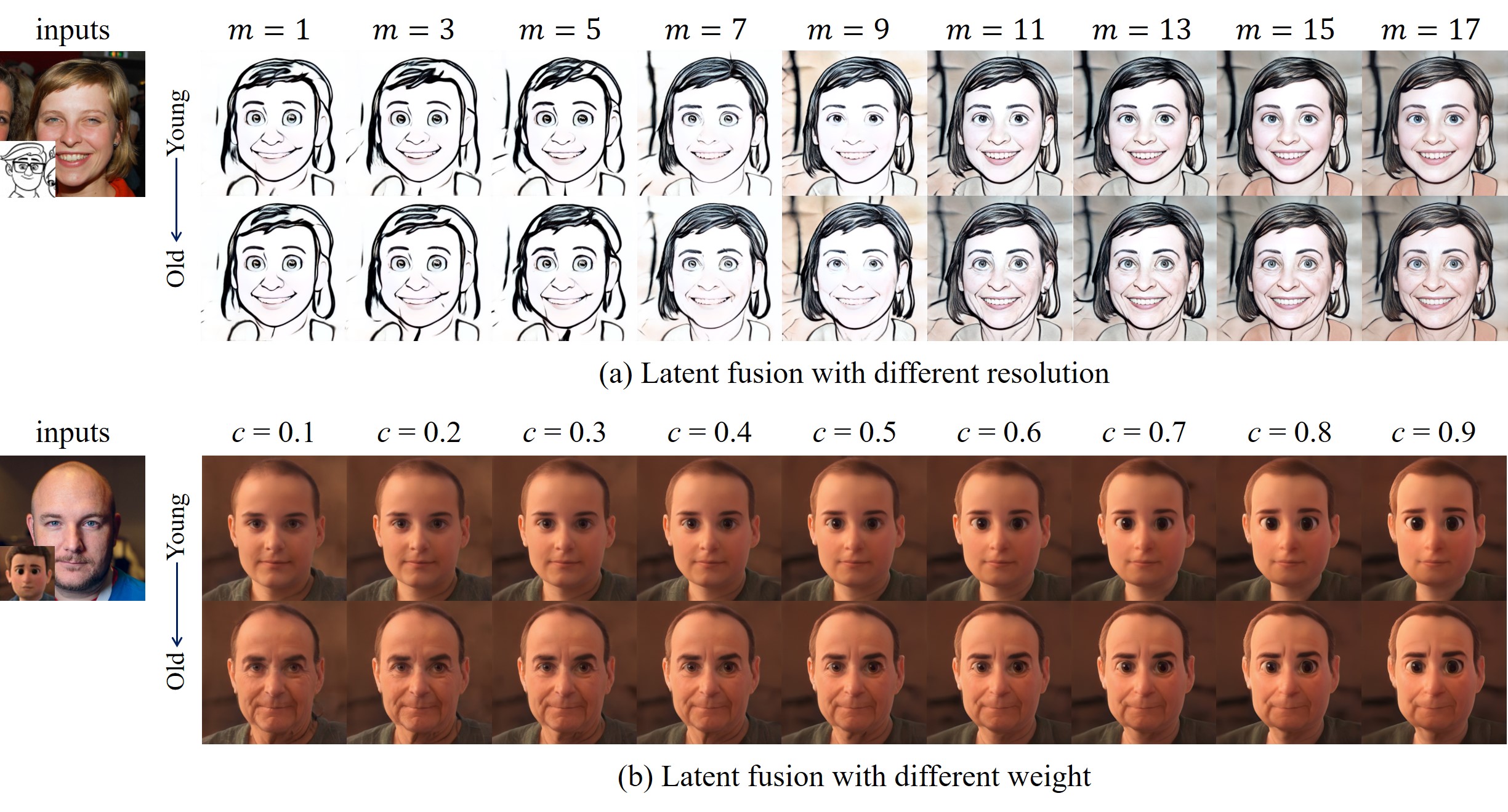}
  \caption{Effect of latent fusion level of \textit{ToonAging}. (a) We vary $m$ from 1 to 17 to observe the re-aging effect while fixing weights as $\textit{\textbf{c}}_{1:m} = \textbf{0.5}$ and $\textit{\textbf{s}}_{m+1:18} = \textbf{1}$, (b) We vary the value $\textit{c}$ from 0.1 to 0.9 to control the re-aging level in the same convolutions while fixing $m=7$.}
  \label{fig:latent_resolution_analysis}
\end{figure*}

\subsection{Preliminaries} \label{method:01}%

\subsubsection{Face Re-Aging Network} \label{pre_sam}

Face re-aging semantically generates a re-aging effect based on the original appearance, following the target age. Many works \cite{hsu2022agetransgan, gomez2022cusp, makhmudkhujaev2021ragan, yao2021hrfae, or2020lifespan, makhmudkhujaev2023raganpp} have proposed re-aging networks for high-quality and dramatic age progression. SAM \cite{alaluf2021only}, a method based on StyleGAN, demonstrates notable effectiveness in encoding age-related attributes into the latent vector, utilizing residuals in the $\mathcal{W+}$ space in conjunction with the StyleGAN and inversion scheme. Its integration capabilities and performance make it a prominent choice among various face re-aging networks. Technically, given an input image $x$, an inversion encoder $E_{\text{inv}}$, and a target age $a_{\text{target}}$, the face re-aging network SAM is formulated as:

\begin{equation} \label{eq:sam}
    \textit{SAM}(x, a_{\text{target}}) := G_{\text{org}}(w^+_{\text{age}}),
\end{equation}

\noindent where $G_{\text{org}}$ denotes the StyleGAN generator, and $w^+_{\text{age}} \in \mathbb{R}^{18 \times 512}$ represents the age-included latent vector encoded as $w^+_{\text{age}} = E_{\text{age}}(x, a_{\text{target}}) + E_{\text{inv}}(x)$, with $a_{\text{target}}$ being a constant ranging from 0 to 100. It is noteworthy that SAM only requires an additional encoder, the age encoder $E_{\text{age}}$. Thus, in the absence of the age encoder ($E_{\text{age}}(x, a_{\text{target}}) := \textbf{0}$), Eq. \ref{eq:sam} effectively reverts to the original StyleGAN formulation $G_{\text{org}}(E_{\text{inv}}(x))$, demonstrating the foundational role of the age encoder in extending the model's capabilities for age-specific transformations. This integration of a simple yet impactful age encoder into the latent vector-based re-aging process underscores our rationale for selecting this model as the basis for our \textit{ToonAging} method.

\subsubsection{Artistic Portrait Style Transfer Network} \label{pre_dualstylegan}

DualStyleGAN stands out among recent advancements \cite{pinkney2020resolution,kim2022cross,chong2022jojogan,back2021fine} in face style transfer networks due to its innovative structure incorporating both intrinsic and extrinsic paths. This design simplifies the style transfer process, effectively integrates different latent vectors, and excels in applying diverse external styles while maintaining core features. Its streamlined yet versatile nature makes it an ideal choice for our methodology, particularly in scenarios requiring the application of diverse external styles. Given PR input $x$ and NPR style $S$, DualStyleGAN is formulated as:

\begin{equation} \label{eq:dualstylegan}
    \textit{DualStyleGAN}(x, S) := G_{\text{modified}}(f_{\text{in}}(z^+_{\text{in}}), f_{\text{ex}}(z^+_{\text{ex}}), \textbf{w}),
\end{equation}

\noindent where $z^+_{\text{in}}$ is the latent vector encoded from PR image via intrinsic path as $z^+_{\text{in}} = E_{\text{inv}}(x) \in \mathcal{Z}+$ \cite{song2021agilegan}, 
$z^+_{\text{ex}}$ is the style-related latent vector encoded from the style via the extrinsic path as $z^+_{\text{ex}} = E_{\text{inv}}(S) \in \mathcal{Z}+$, $\textbf{w}$ is the control weight to determine the style level, and $G_{\text{modified}}$ includes the original generator $G_{\text{org}}$ and the proposed extrinsic style path, which also consists of an inversion network \cite{richardson2021psp}. Through Eq. \ref{eq:sam} and Eq. \ref{eq:dualstylegan}, we found that all the parts are equally used except $\textbf{w}$ which is for style transfer. From that, we would like to design our approach by leveraging the advantages of each network while retaining the frozen parts. To validate the feasibility of this combination and gain a better understanding, we analyze the behavior of each latent vector in the context of face re-aging and portrait style transfer, as described in the methods in Sec. \ref{pre_sam} and Sec. \ref{pre_dualstylegan}.

\subsection{ToonAging}\label{method:02}

In our study, the goal is to perform re-aging and portrait style transfer simultaneously. To achieve this, we first explore the latent vector behavior based on the methods described in Sec. \ref{pre_sam} and Sec. \ref{pre_dualstylegan}, using SAM for face re-aging and DualStyleGAN for portrait style transfer, respectively. To examine face re-aging, we freeze the fine-level elements to highlight the coarse-level effects in age-related latent vector $w^+_{\text{age}} \in \mathbb{R}^{18 \times 512}$ in Eq. \ref{eq:sam}. Additionally, we conduct the reverse experiment by freezing the coarse-level elements and allowing the fine-level elements to change, in order to observe the fine-level effects on the age-related latent vector $w^+{\text{age}}$. From Fig. \ref{fig:simulation_sam}, we found that the model without fine-level can generate the perceptually dominant point for re-aging. Similarly, we exclude the extrinsic style as above. It means $f_{\text{ex}}(z^+_{\text{ex}})$ (Eq. \ref{eq:dualstylegan}) has no effects at the coarse-level or fine-level in each analysis. Following DualStyleGAN \cite{yang2022pastiche}, we also confirmed that elements in the fine-level layer manage the color. In the coarse-level, since we set $\textbf{w} = \textbf{0.5}$, the model used a mixed one for structure information by fusing the intrinsic style $f_{\text{ex}}(z^+_{\text{ex}})$ (Eq. \ref{eq:dualstylegan}). From this analysis, we expect that the fusion on the intrinsic path can accommodate the re-aging effects while preserving the extrinsic path.

\begin{figure}[t]
  \centering
    \includegraphics[width=0.95\linewidth]{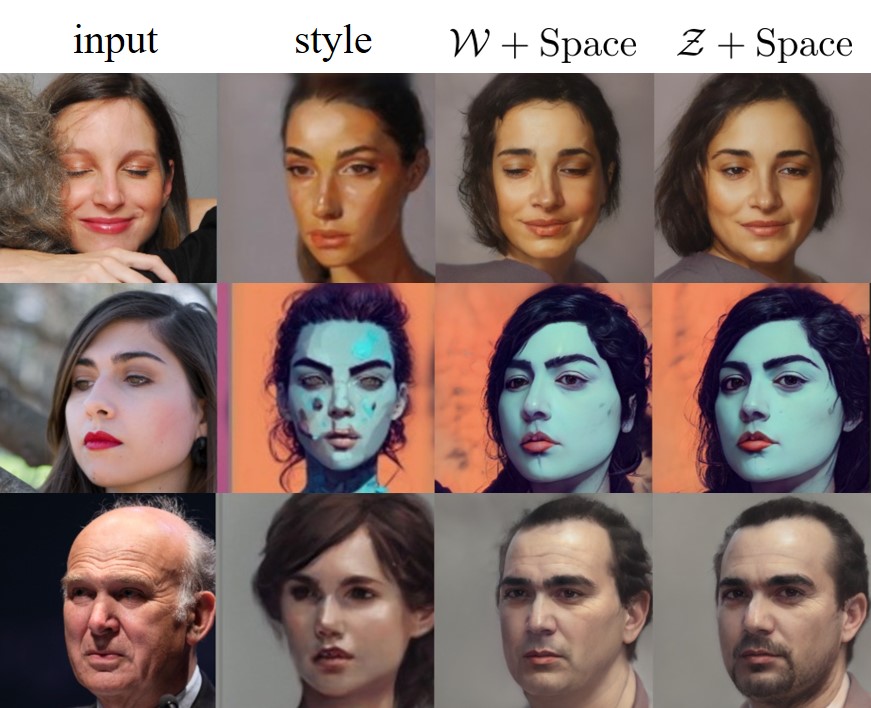}
  \caption{Latent space comparison for intrinsic style encoding. We found that the $\mathcal{W}+$ space can better preserve the facial attributes of the input image (\textit{e.g.}, eye gazing, expression) than the $\mathcal{Z}+$ space.}
  \label{fig:wz_space}
\end{figure}

 \begin{figure*}[!t]
  \centering
    \includegraphics[width=0.95\linewidth]{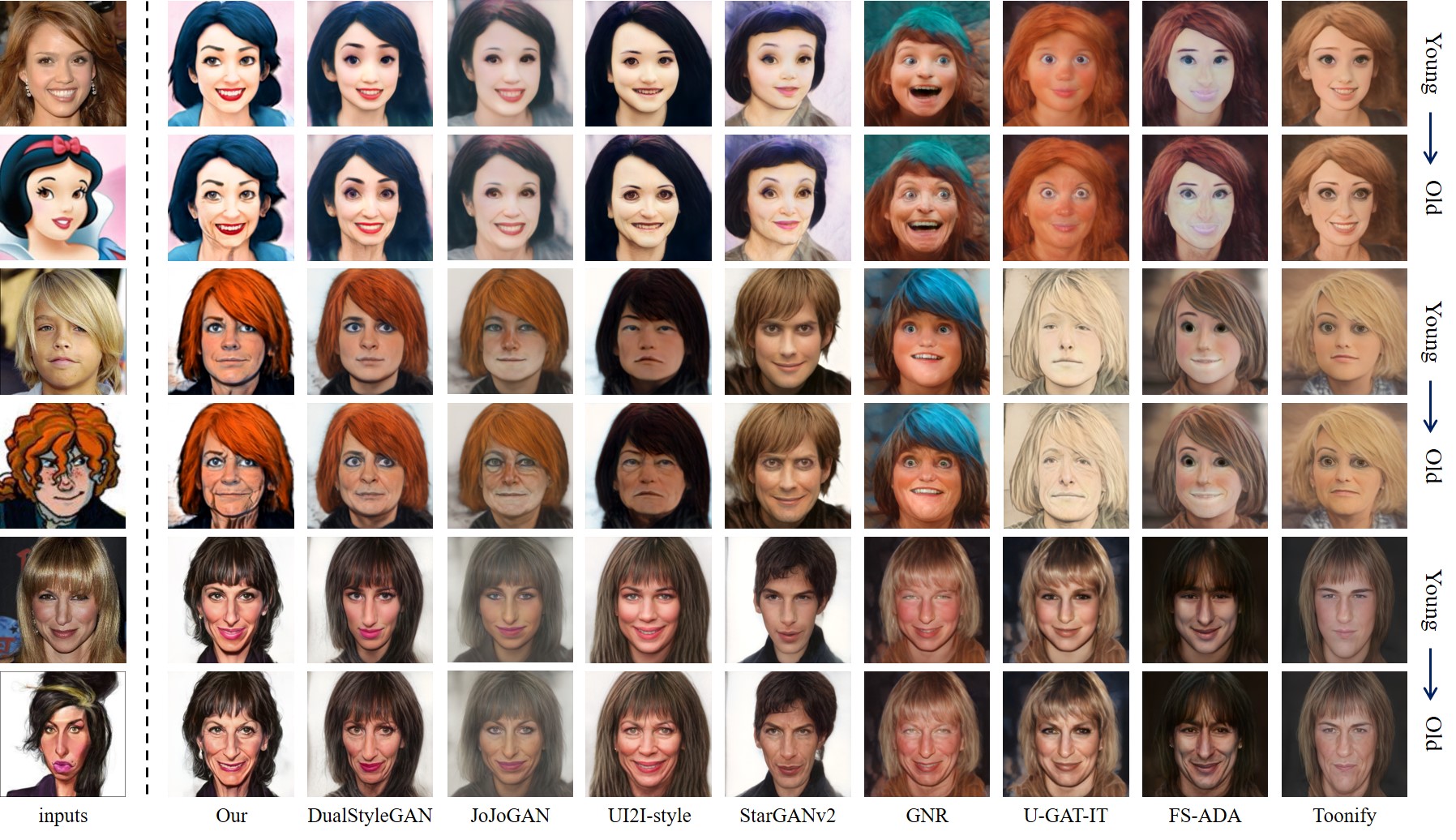}
  \caption{Qualitative comparison with various style transfer networks using a fixed re-aging network, SAM.}
\label{fig:qual_comparison}
\end{figure*}
Hence, we replace the intrinsic style path $z^+_{\text{in}}$ in DualStyleGAN with the SAM encoded style $w^+{\text{SAM}}$ without crossing the $\mathcal{Z}+$ space to use the $\mathcal{W}+$ space directly, while preserving the extrinsic style path $z^+_{\text{ex}}$ in $\mathcal{Z}+$. This is because we found that $\mathcal{W}+$ showed more faithful results in terms of attribute preservation, unlike DualStyleGAN, which uses the $\mathcal{Z}+$ space for the intrinsic path \cite{yang2022pastiche}, as shown in Fig. \ref{fig:wz_space}. By doing that, we can accommodate the $a_{\text{target}}$ and $S$ for face re-aging and exemplar-based portrait style transfer, respectively. The layer swapping-based approach \cite{pinkney2020resolution, kim2022cross, kwak2022generate} has a limitation in that a trained network is required for each domain, which reduces flexibility. Thus, our \textit{ToonAging} is formulated as:    

\begin{equation} \label{eq:toonaging}
    \textit{ToonAging}(x, a_{\text{target}}, S) := G_{\text{modified}}(w^+_{\text{age}}, f_{\text{ex}}(z^+_{\text{ex}}), \textbf{w})
\end{equation}

\noindent \textbf{w} $\in$ $\mathbb{R}^{18}$ is a control weight used to combine the two latent vectors, with the default value set as \textbf{1}, as used in \cite{yang2022pastiche} (Eq. \ref{eq:dualstylegan}). By doing so, we can incorporate the re-aging latent code, which manages age-related attributes, into the cartoonization process. Unlike DualStyleGAN, which embeds real faces into the $\mathcal{Z}+$ space, our approach utilizes the $\mathcal{W}+$ space to directly input the age vector $w^+_{\text{age}}$, without additional processing. In terms of the generator, the final input latent vector $w^+_{\text{ToonAging}}$ becomes $w^+_{\text{ToonAging}} = w^+_{\text{age}} \times \textbf{w} + f_{\text{ex}}(z^+_{\text{ex}}) \times (1 - \textbf{w})$ from $64^2$ to $1024^2$, and it is used in the same way as DualStyleGAN formulation (read the details in \cite{yang2022pastiche}) for the coarse-level ($4^2$ to $32^2$). All the encoders are based on \textit{pSp} \cite{richardson2021psp}. Our scheme is illustrated in Fig. \ref{fig:toonaing_architecture}.

In summary, the NPR face image is generated from the extrinsic style path \cite{yang2022pastiche}, with the chosen style $S$ from the exemplar database. Re-aging effects are represented by following the target age $a_{\text{target}}$ via the age vector $w^+_{\text{age}}$ through SAM encoding \cite{alaluf2021only}. By default, input vectors $(w_{\text{in}}, w_{\text{ex}})$ are set as $(w^+_{\text{age}}, f_{\text{ex}}(z^+_{\text{ex}}))$ in Eq. \ref{eq:toonaging}, but for random generation it can be $(w_{\text{in}}, w_{\text{ex}}) = (f_{\text{in}}(\textbf{z}_{\text{in}}), f_{\text{ex}}(\textbf{z}_{\text{ex}}))$ where $\textbf{z}_{in}$ and $\textbf{z}_{ex}$ are  generated from gaussian distribution. Following DualStyleGAN \cite{yang2022pastiche}, when $\textbf{w} = \textbf{0}$, $\textit{ToonAging}(x, a_{\text{target}}, S)$ degrades into a standard re-aging generator, \textit{i.e.}, $G_{\text{modified}}(w^+_{\text{age}}, \cdot, \textbf{0}) \sim SAM(x || a_{\text{target}}) := G_{\text{org}}(E_{\text{age}}(x, a_{\text{target}}) + E_{\text{inv}}(x))$ (Eq. \ref{eq:sam}).

\subsection{Latent Fusion Analysis}\label{sec:latentfusionanalysis}

Our main objective with \textit{ToonAging} is to effectively integrate re-aging and style transfer. To achieve this, we delve into the latent vector from coarse to fine levels. Among the 18 different vectors $w \in \mathbb{R}^{18 \times 512}$, we control the resolution level by adjusting the control weight \textbf{w} from the coarse level as follows:

\begin{equation} \label{eq:latent_fusion}
\textbf{w} = \{\textbf{\textit{c}}_{1,\dots,m}\} \cup \{\textbf{\textit{s}}_{m+1,\dots,18}\}
\end{equation} 

\noindent here, $\textbf{\textit{c}}$ represents the re-aging components, while $\textbf{\textit{s}}$ represents the style transfer components in the vector. The parameter $m$ controls the range to constrain the elements from $1$ to $m$. In \textit{ToonAging}, $\textbf{w}$ determines the layer up to which the re-aging effects are applied out of the 18 layers. As illustrated in Fig. \ref{fig:latent_resolution_analysis} (a), we observed that setting $m > 7$ results in a noticeable deviation in the style from $S$, leading to the generation of colors similar to those in input $x$. Moreover, when adjusting the magnitude $c$ of the re-aging latent components, we discovered that values around 0.5 yield credible outcomes, as depicted in Fig. \ref{fig:latent_resolution_analysis} (b). As a result, \textit{ToonAging} defaults to using $m = 7$ and $c = 0.5$. However, both $m$ and $c$ remain adjustable parameters, enabling users to fine-tune these values according to their preferences.

\section{Experiments}

\subsection{Experimental Setup}

 All the network components, including the generator, mapping network, and inversion network, were utilized with pre-trained weights. We used SAM weights from the official Pytorch implementation\footnote{https://github.com/yuval-alaluf/SAM} and the codebase of DualStyleGAN\footnote{https://github.com/williamyang1991/DualStyleGAN}, aligning our methodology with existing frameworks without undergoing a separate training process. Our code is implemented in Pytorch, inheriting the StyleGAN architecture \cite{karras2019stylegan}, which enables the generation of output images up to $1024^2$ resolution. Using the NVIDIA RTX 3090 GPU, we achieve a generation rate of 4 images per second, excluding the time for the extrinsic style optimization process \cite{yang2022pastiche}. We set $\textit{c}_{1,\dots,m} = 0.5$ and $\textbf{\textit{s}}_{m+1,\dots,18} = 1$ with $m = 7$ as default settings. 

\subsection{Qualitative Results}

For a qualitative comparison, in the absence of any existing network capable of performing artistic style transfer and facial re-aging simultaneously, our \textit{ToonAging} technique is assessed against a naive approach that processes style transfer followed by re-aging. In our experiments, we specifically target the ages of 10 and 55 to represent the younger and older age directions, respectively. The input images are obtained from the FFHQ dataset, and the style images, which include cartoons, caricatures, and Pixar-inspired themes, are directly sourced from the images used in the DualStyleGAN figure for comparison. The effectiveness of \textit{ToonAging} in achieving style transfer and age progression is then benchmarked against several established methods, including DualStyleGAN \cite{yang2022pastiche}, JoJoGAN \cite{chong2022jojogan}, U2I-style \cite{huang2021unsupervised}, StarGANv2 \cite{choi2020starganv2}, GNR \cite{chong2021gnr}, U-GAT-IT \cite{kim2019ugatit}, FS-ADA \cite{ojha2021fsada}, and Toonify \cite{pinkney2020resolution}.

As illustrated in Fig. \ref{fig:qual_comparison}, our method excels at delivering natural age progression with accurate skin textures and wrinkles, whereas other techniques may produce less realistic effects and often sacrifice the unique features of style transfer in an attempt to create images that resemble real people. Methods like DualStyleGAN \cite{yang2022pastiche}, JoJoGAN \cite{chong2022jojogan}, and UI2I-style \cite{huang2021unsupervised} can struggle with maintaining age-related authenticity while preserving the distinctive characteristics of the original style. Similarly, StarGANv2 \cite{choi2020starganv2} may distort facial features, while techniques such as GNR \cite{chong2021gnr}, U-GAT-IT \cite{kim2019ugatit}, FS-ADA \cite{ojha2021fsada}, and Toonify \cite{pinkney2020resolution} each have their own limitations, ranging from loss of detail to an inability to fully capture the essence of aging. In contrast, our method consistently offers more lifelike and coherent age transformation results without compromising the stylistic elements of the original image.

\begin{figure*}[t]
  \centering  
  \includegraphics[width=0.95\textwidth]{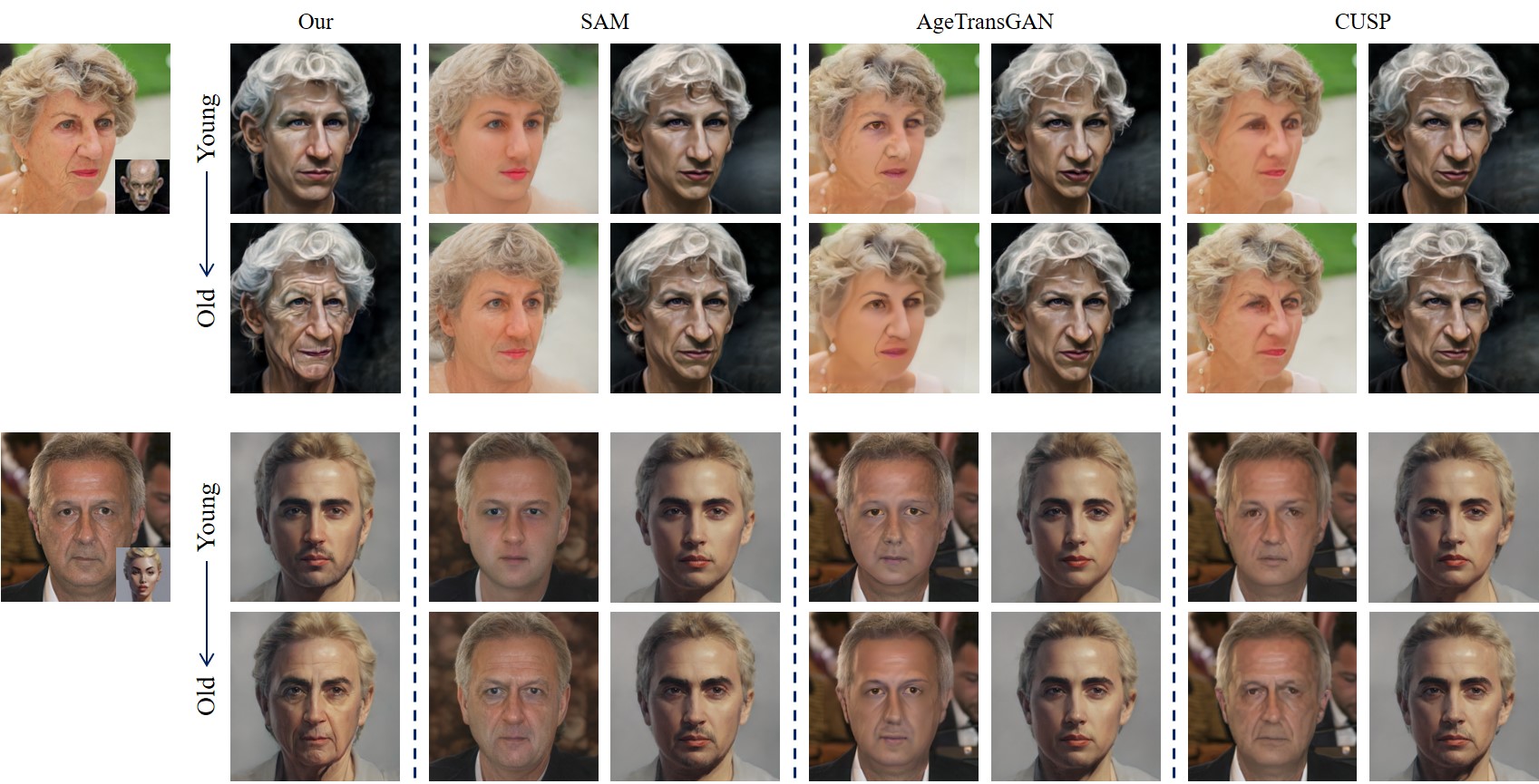}
  \caption{Qualitative comparison with various face re-aging networks, including SAM \cite{alaluf2021only}, CUSP \cite{gomez2022cusp}, and AgeTransGAN \cite{hsu2022agetransgan}, while fixing the style transfer model as DualStyleGAN \cite{yang2022pastiche}. Our \textit{ToonAging} method outperforms all the other approaches, as these approaches fail to conserve aging-related visual effects through a 2-stage process.
  }
  \label{fig:naive_approach_naive1}
\end{figure*}

In the naive approach, we experiment with various state-of-the-art image-level face re-aging networks, including SAM \cite{alaluf2021only}, AgeTransGAN \cite{hsu2022agetransgan}, and CUSP \cite{gomez2022cusp}. As demonstrated in Fig. \ref{fig:naive_approach_naive1}, our \textit{ToonAging} method outperforms all the naive approaches utilizing various state-of-the-art re-aging models. Here, we employ SAM as the re-aging component in \textit{ToonAging}, which generates more natural aging effects compared to the other models, as evidenced in Fig. \ref{fig:naive_approach_naive1}.

\begin{table}[b]
\centering
\caption{User preference scores. Best scores are marked in \textbf{bold}.}
\begin{tabular}{@{}lccc@{}}
\toprule
 Domain & SR & RS & \textbf{Ours} \\
\midrule
\textit{Arcane} & 0.2 & 0.26 & \textbf{0.54} \\
\textit{Caricature} & 0.32 & 0.22 & \textbf{0.46} \\
\textit{Cartoon} & 0.22 & 0.31 & \textbf{0.46} \\
\textit{Illustration} & 0.24 & 0.27 & \textbf{0.49} \\
\textit{Impasto} & 0.22 & 0.25 & \textbf{0.53} \\
\textit{Pixar} & 0.23 & 0.22 & \textbf{0.54} \\
\bottomrule
\end{tabular}
\label{tab:user_study}
\end{table}

\subsection{Quantitative Results}

To address the challenges of accurately estimating age and identity in NPR images and the lack of standard metrics for evaluating face re-aging in this domain, we conducted a comprehensive user study as presented in Tab. \ref{tab:user_study}. The study included a diverse set of styles from six domains: \textit{Arcane}, \textit{Caricature}, \textit{Cartoon}, \textit{Illustration}, \textit{Impasto}, and \textit{Pixar}, each with 10 samples, totaling 60 samples. Participants were tasked with identifying the images that most effectively conveyed the intended age transformation and style adaptation, allowing us to assess user preferences and the perceived success of the transformations. 

The study yielded 589 responses, demonstrating high engagement and providing valuable insights into user preferences. Our proposed method, \textit{ToonAging}, consistently outperformed other approaches, achieving the highest scores in all categories. This result validates the effectiveness of \textit{ToonAging} in generating convincing and visually appealing re-aged representations within the NPR context. Furthermore, the outcome highlights the importance of incorporating user-centric evaluations to complement traditional metrics and gain a more comprehensive understanding of face re-aging performance in the NPR domain.

\section{Conclusion}

In this paper, we present \textit{ToonAging}, which focuses on combining the dual challenges of face re-aging and portrait style transfer. Since the naive two-stage approach lacks the capability to represent both re-aging visuals and style transfer effectively, we devised a one-stage method that leverages the strengths of existing networks. Our \textit{ToonAging} builds upon portrait style transfer, incorporating a face re-aging network and demonstrating plausible performance with enhanced editability for stylization and face re-aging. In our experiments, \textit{ToonAging} outperformed existing methods in terms of both naturalness and image quality. \textit{ToonAging} serves as a versatile tool for human face creation, with potential applications in various NPR face manipulation tasks involving face re-aging and style transfer.

\section*{Acknowledgement}
This research was supported by Culture, Sports and Tourism R\&D Program through the Korea Creative Content Agency(KOCCA) grant funded by the Ministry of Culture, Sports and Tourism(MCST) in 2023 (Project Name: Development of digital abusing detection and management technology for a safe Metaverse service, Project Number: RS-2023-00227686, Contribution Rate: 50\%) and the National Research Foundation of Korea (NRF) grant funded by the Korean government (MSIT) (No.2022R1A2C1004657, Contribution Rate: 50\%).

{\small
\bibliographystyle{ieee_fullname}
\bibliography{main}

\begin{thebibliography}{10}\itemsep=-1pt

\bibitem{alaluf2021only}
Yuval Alaluf, Or Patashnik, and Daniel Cohen-Or.
\newblock Only a matter of style: Age transformation using a style-based regression model.
\newblock {\em ACM Transactions on Graphics (TOG)}, 40(4):1--12, 2021.

\bibitem{back2021fine}
Jihye Back.
\newblock Fine-tuning stylegan2 for cartoon face generation.
\newblock {\em arXiv preprint arXiv:2106.12445}, 2021.

\bibitem{back2022webtoonme}
Jihye Back, Seungkwon Kim, and Namhyuk Ahn.
\newblock Webtoonme: A data-centric approach for full-body portrait stylization.
\newblock In {\em SIGGRAPH Asia 2022 Technical Communications}, pages 1--4. ACM, 2022.

\bibitem{chefer2022image}
Hila Chefer, Sagie Benaim, Roni Paiss, and Lior Wolf.
\newblock Image-based clip-guided essence transfer.
\newblock In {\em European Conference on Computer Vision}, pages 695--711. Springer, 2022.

\bibitem{chen2023fading}
Xiangyi Chen and St{\'e}phane Lathuili{\`e}re.
\newblock Face aging via diffusion-based editing.
\newblock {\em arXiv preprint arXiv:2309.11321}, 2023.

\bibitem{choi2020starganv2}
Yunjey Choi, Youngjung Uh, Jaejun Yoo, and Jung-Woo Ha.
\newblock Stargan v2: Diverse image synthesis for multiple domains.
\newblock In {\em Proceedings of the IEEE/CVF conference on computer vision and pattern recognition}, pages 8188--8197, 2020.

\bibitem{chong2021gnr}
Min~Jin Chong and David Forsyth.
\newblock Gans n'roses: Stable, controllable, diverse image to image translation (works for videos too!).
\newblock {\em arXiv preprint arXiv:2106.06561}, 2021.

\bibitem{chong2022jojogan}
Min~Jin Chong and David Forsyth.
\newblock Jojogan: One shot face stylization.
\newblock In {\em European Conference on Computer Vision}, pages 128--152. Springer, 2022.

\bibitem{duong2019automatic}
Chi~Nhan Duong, Khoa Luu, Kha~Gia Quach, Nghia Nguyen, Eric Patterson, Tien~D Bui, and Ngan Le.
\newblock Automatic face aging in videos via deep reinforcement learning.
\newblock In {\em Proceedings of the IEEE/CVF Conference on Computer Vision and Pattern Recognition}, pages 10013--10022, 2019.

\bibitem{gomez2022cusp}
Guillermo Gomez-Trenado, St{\'e}phane Lathuili{\`e}re, Pablo Mesejo, and {\'O}scar Cord{\'o}n.
\newblock Custom structure preservation in face aging.
\newblock In {\em European Conference on Computer Vision}, pages 565--580. Springer, 2022.

\bibitem{hsu2022agetransgan}
Gee-Sern Hsu, Rui-Cang Xie, Zhi-Ting Chen, and Yu-Hong Lin.
\newblock Agetransgan for facial age transformation with rectified performance metrics.
\newblock In {\em European Conference on Computer Vision}, pages 580--595. Springer, 2022.

\bibitem{huang2021unsupervised}
Jialu Huang, Jing Liao, and Sam Kwong.
\newblock Unsupervised image-to-image translation via pre-trained stylegan2 network.
\newblock {\em IEEE Transactions on Multimedia}, 24:1435--1448, 2021.

\bibitem{huang2018munit}
Xun Huang, Ming-Yu Liu, Serge Belongie, and Jan Kautz.
\newblock Multimodal unsupervised image-to-image translation.
\newblock In {\em Proceedings of the European conference on computer vision (ECCV)}, pages 172--189, 2018.

\bibitem{jang2021stylecarigan}
Wonjong Jang, Gwangjin Ju, Yucheol Jung, Jiaolong Yang, Xin Tong, and Seungyong Lee.
\newblock Stylecarigan: caricature generation via stylegan feature map modulation.
\newblock {\em ACM Transactions on Graphics (TOG)}, 40(4):1--16, 2021.

\bibitem{karras2020ada}
Tero Karras, Miika Aittala, Janne Hellsten, Samuli Laine, Jaakko Lehtinen, and Timo Aila.
\newblock Training generative adversarial networks with limited data.
\newblock {\em Advances in neural information processing systems}, 33:12104--12114, 2020.

\bibitem{karras2019stylegan}
Tero Karras, Samuli Laine, and Timo Aila.
\newblock A style-based generator architecture for generative adversarial networks.
\newblock In {\em Proceedings of the IEEE/CVF conference on computer vision and pattern recognition}, pages 4401--4410, 2019.

\bibitem{karras2020stylegan2}
Tero Karras, Samuli Laine, Miika Aittala, Janne Hellsten, Jaakko Lehtinen, and Timo Aila.
\newblock Analyzing and improving the image quality of stylegan.
\newblock In {\em Proceedings of the IEEE/CVF conference on computer vision and pattern recognition}, pages 8110--8119, 2020.

\bibitem{khowaja2023face}
Sunder~Ali Khowaja, Ghulam Mujtaba, Jiseok Yoon, and Ik~Hyun Lee.
\newblock Face-past: Facial pose awareness and style transfer networks.
\newblock {\em arXiv preprint arXiv:2307.09020}, 2023.

\bibitem{kim2024minecraft}
Bumsoo Kim, Sanghyun Byun, Yonghoon Jung, Wonseop Shin, Sareer~UI Amin, and Sanghyun Seo.
\newblock Minecraft-ify: Minecraft style image generation with text-guided image editing for in-game application.
\newblock {\em arXiv preprint arXiv:2402.05448}, 2024.

\bibitem{kim2023context}
Doyeon Kim, Eunji Ko, Hyunsu Kim, Yunji Kim, Junho Kim, Dongchan Min, Junmo Kim, and Sung~Ju Hwang.
\newblock Context-preserving two-stage video domain translation for portrait stylization.
\newblock {\em arXiv preprint arXiv:2305.19135}, 2023.

\bibitem{kim2023diffusionvae}
Gyeongman Kim, Hajin Shim, Hyunsu Kim, Yunjey Choi, Junho Kim, and Eunho Yang.
\newblock Diffusion video autoencoders: Toward temporally consistent face video editing via disentangled video encoding.
\newblock In {\em Proceedings of the IEEE/CVF Conference on Computer Vision and Pattern Recognition}, pages 6091--6100, 2023.

\bibitem{kim2019u}
Junho Kim, Minjae Kim, Hyeonwoo Kang, and Kwanghee Lee.
\newblock U-gat-it: Unsupervised generative attentional networks with adaptive layer-instance normalization for image-to-image translation.
\newblock {\em arXiv preprint arXiv:1907.10830}, 2019.

\bibitem{kim2019ugatit}
Junho Kim, Minjae Kim, Hyeonwoo Kang, and Kwanghee Lee.
\newblock U-gat-it: Unsupervised generative attentional networks with adaptive layer-instance normalization for image-to-image translation.
\newblock {\em arXiv preprint arXiv:1907.10830}, 2019.

\bibitem{kim2022cross}
Seungkwon Kim, Chaeheon Gwak, Dohyun Kim, Kwangho Lee, Jihye Back, Namhyuk Ahn, and Daesik Kim.
\newblock Cross-domain style mixing for face cartoonization.
\newblock {\em arXiv preprint arXiv:2205.12450}, 2022.

\bibitem{kwak2022generate}
Jeong-gi Kwak, Yuanming Li, Dongsik Yoon, David Han, and Hanseok Ko.
\newblock Generate and edit your own character in a canonical view.
\newblock {\em arXiv preprint arXiv:2205.02974}, 2022.

\bibitem{lee2023fix}
Dongyeun Lee, Jae~Young Lee, Doyeon Kim, Jaehyun Choi, Jaejun Yoo, and Junmo Kim.
\newblock Fix the noise: Disentangling source feature for controllable domain translation.
\newblock In {\em Proceedings of the IEEE/CVF Conference on Computer Vision and Pattern Recognition}, pages 14224--14234, 2023.

\bibitem{lee2018drit}
Hsin-Ying Lee, Hung-Yu Tseng, Jia-Bin Huang, Maneesh Singh, and Ming-Hsuan Yang.
\newblock Diverse image-to-image translation via disentangled representations.
\newblock In {\em Proceedings of the European conference on computer vision (ECCV)}, pages 35--51, 2018.

\bibitem{lee2020dritpp}
Hsin-Ying Lee, Hung-Yu Tseng, Qi Mao, Jia-Bin Huang, Yu-Ding Lu, Maneesh Singh, and Ming-Hsuan Yang.
\newblock Drit++: Diverse image-to-image translation via disentangled representations.
\newblock {\em International Journal of Computer Vision}, 128:2402--2417, 2020.

\bibitem{li2023multimodal}
Mengtian Li, Yi Dong, Minxuan Lin, Haibin Huang, Pengfei Wan, and Chongyang Ma.
\newblock Multi-modal face stylization with a generative prior.
\newblock {\em arXiv preprint arXiv:2305.18009}, 2023.

\bibitem{li2023pluralistic}
Peipei Li, Rui Wang, Huaibo Huang, Ran He, and Zhaofeng He.
\newblock Pluralistic aging diffusion autoencoder.
\newblock In {\em Proceedings of the IEEE/CVF International Conference on Computer Vision}, pages 22613--22623, 2023.

\bibitem{li2023parsing}
Zhansheng Li, Yangyang Xu, Nanxuan Zhao, Yang Zhou, Yongtuo Liu, Dahua Lin, and Shengfeng He.
\newblock Parsing-conditioned anime translation: A new dataset and method.
\newblock {\em ACM Transactions on Graphics}, 42(3):1--14, 2023.

\bibitem{liu2023portrait}
Jin Liu, Huaibo Huang, Chao Jin, and Ran He.
\newblock Portrait diffusion: Training-free face stylization with chain-of-painting.
\newblock {\em arXiv preprint arXiv:2312.02212}, 2023.

\bibitem{liu2017unit}
Ming-Yu Liu, Thomas Breuel, and Jan Kautz.
\newblock Unsupervised image-to-image translation networks.
\newblock {\em Advances in neural information processing systems}, 30, 2017.

\bibitem{maeng2023gmba}
Junyeong Maeng, Kwanseok Oh, and Heung-Il Suk.
\newblock Age-aware guidance via masking-based attention in face aging.
\newblock In {\em Proceedings of the 32nd ACM International Conference on Information and Knowledge Management}, pages 4165--4169, 2023.

\bibitem{makhmudkhujaev2021ragan}
Farkhod Makhmudkhujaev, Sungeun Hong, and In~Kyu Park.
\newblock Re-aging gan: Toward personalized face age transformation.
\newblock In {\em Proceedings of the IEEE/CVF International Conference on Computer Vision}, pages 3908--3917, 2021.

\bibitem{makhmudkhujaev2023raganpp}
Farkhod Makhmudkhujaev, Sungeun Hong, and In~Kyu Park.
\newblock Re-aging gan++: Temporally consistent transformation of faces in videos.
\newblock {\em IEEE Access}, 11:137377--137386, 2023.

\bibitem{men2022dct}
Yifang Men, Yuan Yao, Miaomiao Cui, Zhouhui Lian, and Xuansong Xie.
\newblock Dct-net: domain-calibrated translation for portrait stylization.
\newblock {\em ACM Transactions on Graphics (TOG)}, 41(4):1--9, 2022.

\bibitem{men2022unpaired}
Yifang Men, Yuan Yao, Miaomiao Cui, Zhouhui Lian, Xuansong Xie, and Xian-Sheng Hua.
\newblock Unpaired cartoon image synthesis via gated cycle mapping.
\newblock In {\em Proceedings of the IEEE/CVF Conference on Computer Vision and Pattern Recognition}, pages 3501--3510, 2022.

\bibitem{muqeet2023video}
Abdul Muqeet, Kyuchul Lee, Bumsoo Kim, Yohan Hong, Hyungrae Lee, Woonggon Kim, and KwangHee Lee.
\newblock Video face re-aging: Toward temporally consistent face re-aging.
\newblock {\em arXiv preprint arXiv:2311.11642}, 2023.

\bibitem{ojha2021fsada}
Utkarsh Ojha, Yijun Li, Jingwan Lu, Alexei~A Efros, Yong~Jae Lee, Eli Shechtman, and Richard Zhang.
\newblock Few-shot image generation via cross-domain correspondence.
\newblock In {\em Proceedings of the IEEE/CVF Conference on Computer Vision and Pattern Recognition}, pages 10743--10752, 2021.

\bibitem{or2020lifespan}
Roy Or-El, Soumyadip Sengupta, Ohad Fried, Eli Shechtman, and Ira Kemelmacher-Shlizerman.
\newblock Lifespan age transformation synthesis.
\newblock In {\em Computer Vision--ECCV 2020: 16th European Conference, Glasgow, UK, August 23--28, 2020, Proceedings, Part VI 16}, pages 739--755. Springer, 2020.

\bibitem{pinkney2020resolution}
Justin~NM Pinkney and Doron Adler.
\newblock Resolution dependent gan interpolation for controllable image synthesis between domains.
\newblock {\em arXiv preprint arXiv:2010.05334}, 2020.

\bibitem{richardson2021psp}
Elad Richardson, Yuval Alaluf, Or Patashnik, Yotam Nitzan, Yaniv Azar, Stav Shapiro, and Daniel Cohen-Or.
\newblock Encoding in style: a stylegan encoder for image-to-image translation.
\newblock In {\em Proceedings of the IEEE/CVF conference on computer vision and pattern recognition}, pages 2287--2296, 2021.

\bibitem{rothe2015dex}
Rasmus Rothe, Radu Timofte, and Luc Van~Gool.
\newblock Dex: Deep expectation of apparent age from a single image.
\newblock In {\em Proceedings of the IEEE international conference on computer vision workshops}, pages 10--15, 2015.

\bibitem{shah2022multistylegan}
Viraj Shah and Svetlana Lazebnik.
\newblock Multistylegan: Multiple one-shot face stylizations using a single gan.
\newblock {\em arXiv preprint arXiv:2210.04120}, 2022.

\bibitem{song2021agilegan}
Guoxian Song, Linjie Luo, Jing Liu, Wan-Chun Ma, Chunpong Lai, Chuanxia Zheng, and Tat-Jen Cham.
\newblock Agilegan: stylizing portraits by inversion-consistent transfer learning.
\newblock {\em ACM Transactions on Graphics (TOG)}, 40(4):1--13, 2021.

\bibitem{song2023agilegan3d}
Guoxian Song, Hongyi Xu, Jing Liu, Tiancheng Zhi, Yichun Shi, Jianfeng Zhang, Zihang Jiang, Jiashi Feng, Shen Sang, and Linjie Luo.
\newblock Agilegan3d: Few-shot 3d portrait stylization by augmented transfer learning.
\newblock {\em arXiv preprint arXiv:2303.14297}, 2023.

\bibitem{song2024styleganfusion}
Kunpeng Song, Ligong Han, Bingchen Liu, Dimitris Metaxas, and Ahmed Elgammal.
\newblock Stylegan-fusion: Diffusion guided domain adaptation of image generators.
\newblock In {\em Proceedings of the IEEE/CVF Winter Conference on Applications of Computer Vision}, pages 5453--5463, 2024.

\bibitem{tzaban2022stitch}
Rotem Tzaban, Ron Mokady, Rinon Gal, Amit Bermano, and Daniel Cohen-Or.
\newblock Stitch it in time: Gan-based facial editing of real videos.
\newblock In {\em SIGGRAPH Asia 2022 Conference Papers}, pages 1--9, 2022.

\bibitem{wang20223d}
Hao Wang, Guosheng Lin, Steven~CH Hoi, and Chunyan Miao.
\newblock 3d cartoon face generation with controllable expressions from a single gan image.
\newblock {\em arXiv preprint arXiv:2207.14425}, 2022.

\bibitem{wang2022realtime}
Xinrui Wang, Zhuoru Li, Xiao Zhou, Yusuke Iwasawa, and Yutaka Matsuo.
\newblock Realtime fewshot portrait stylization based on geometric alignment.
\newblock {\em arXiv preprint arXiv:2211.15549}, 2022.

\bibitem{wang2022ctlgan}
Yue Wang, Ran Yi, Ying Tai, Chengjie Wang, and Lizhuang Ma.
\newblock Ctlgan: Few-shot artistic portraits generation with contrastive transfer learning.
\newblock {\em arXiv preprint arXiv:2203.08612}, 2022.

\bibitem{wu2021stylespace}
Zongze Wu, Dani Lischinski, and Eli Shechtman.
\newblock Stylespace analysis: Disentangled controls for stylegan image generation.
\newblock In {\em Proceedings of the IEEE/CVF Conference on Computer Vision and Pattern Recognition}, pages 12863--12872, 2021.

\bibitem{xiao2022appearance}
Wenpeng Xiao, Cheng Xu, Jiajie Mai, Xuemiao Xu, Yue Li, Chengze Li, Xueting Liu, and Shengfeng He.
\newblock Appearance-preserved portrait-to-anime translation via proxy-guided domain adaptation.
\newblock {\em IEEE Transactions on Visualization and Computer Graphics}, 2022.

\bibitem{xie2022vfhq}
Liangbin Xie, Xintao Wang, Honglun Zhang, Chao Dong, and Ying Shan.
\newblock Vfhq: A high-quality dataset and benchmark for video face super-resolution.
\newblock In {\em Proceedings of the IEEE/CVF Conference on Computer Vision and Pattern Recognition}, pages 657--666, 2022.

\bibitem{yang2022pastiche}
Shuai Yang, Liming Jiang, Ziwei Liu, and Chen~Change Loy.
\newblock Pastiche master: Exemplar-based high-resolution portrait style transfer.
\newblock In {\em Proceedings of the IEEE/CVF Conference on Computer Vision and Pattern Recognition}, pages 7693--7702, 2022.

\bibitem{yang2022vtoonify}
Shuai Yang, Liming Jiang, Ziwei Liu, and Chen~Change Loy.
\newblock Vtoonify: Controllable high-resolution portrait video style transfer.
\newblock {\em ACM Transactions on Graphics (TOG)}, 41(6):1--15, 2022.

\bibitem{yang2023styleganex}
Shuai Yang, Liming Jiang, Ziwei Liu, and Chen~Change Loy.
\newblock Styleganex: Stylegan-based manipulation beyond cropped aligned faces.
\newblock {\em arXiv preprint arXiv:2303.06146}, 2023.

\bibitem{yao2021hrfae}
Xu Yao, Gilles Puy, Alasdair Newson, Yann Gousseau, and Pierre Hellier.
\newblock High resolution face age editing.
\newblock In {\em 2020 25th International conference on pattern recognition (ICPR)}, pages 8624--8631. IEEE, 2021.

\bibitem{yoon2023manipulation}
Dongsik Yoon, Jineui Kim, Vincent Lorant, and Sungku Kang.
\newblock Manipulation of age variation using stylegan inversion and fine-tuning.
\newblock {\em IEEE Access}, 11:131475--131486, 2023.

\bibitem{zhai2018identity}
Zhonghua Zhai and Jian Zhai.
\newblock Identity-preserving conditional generative adversarial network.
\newblock In {\em 2018 International Joint Conference on Neural Networks (IJCNN)}, pages 1--5. IEEE, 2018.

\bibitem{zhang2022generalizedOD}
Zicheng Zhang, Yinglu Liu, Congying Han, Tiande Guo, Ting Yao, and Tao Mei.
\newblock Generalized one-shot domain adaptation of generative adversarial networks.
\newblock {\em Advances in Neural Information Processing Systems}, 35:13718--13730, 2022.

\bibitem{zhou2022hrinversion}
Peng Zhou, Lingxi Xie, Bingbing Ni, Lin Liu, and Qi Tian.
\newblock Hrinversion: High-resolution gan inversion for cross-domain image synthesis.
\newblock {\em IEEE Transactions on Circuits and Systems for Video Technology}, 2022.

\bibitem{zhu2017cyclegan}
Jun-Yan Zhu, Taesung Park, Phillip Isola, and Alexei~A Efros.
\newblock Unpaired image-to-image translation using cycle-consistent adversarial networks.
\newblock In {\em Proceedings of the IEEE international conference on computer vision}, pages 2223--2232, 2017.

\bibitem{zhu2023few}
Runchuan Zhu, Naye Ji, Youbing Zhao, and Fan Zhang.
\newblock Few-shots portrait generation with style enhancement and identity preservation.
\newblock {\em arXiv preprint arXiv:2303.00377}, 2023.

\bibitem{zoss2022production}
Gaspard Zoss, Prashanth Chandran, Eftychios Sifakis, Markus Gross, Paulo Gotardo, and Derek Bradley.
\newblock Production-ready face re-aging for visual effects.
\newblock {\em ACM Transactions on Graphics (TOG)}, 41(6):1--12, 2022.

\end{thebibliography}
}

\clearpage
\appendix
\section{Appendix}

This appendix presents additional experiments.

\subsection{Style-Age Interpolation}

\begin{figure}[!b]
  \centering
  \includegraphics[width=1.0\linewidth]{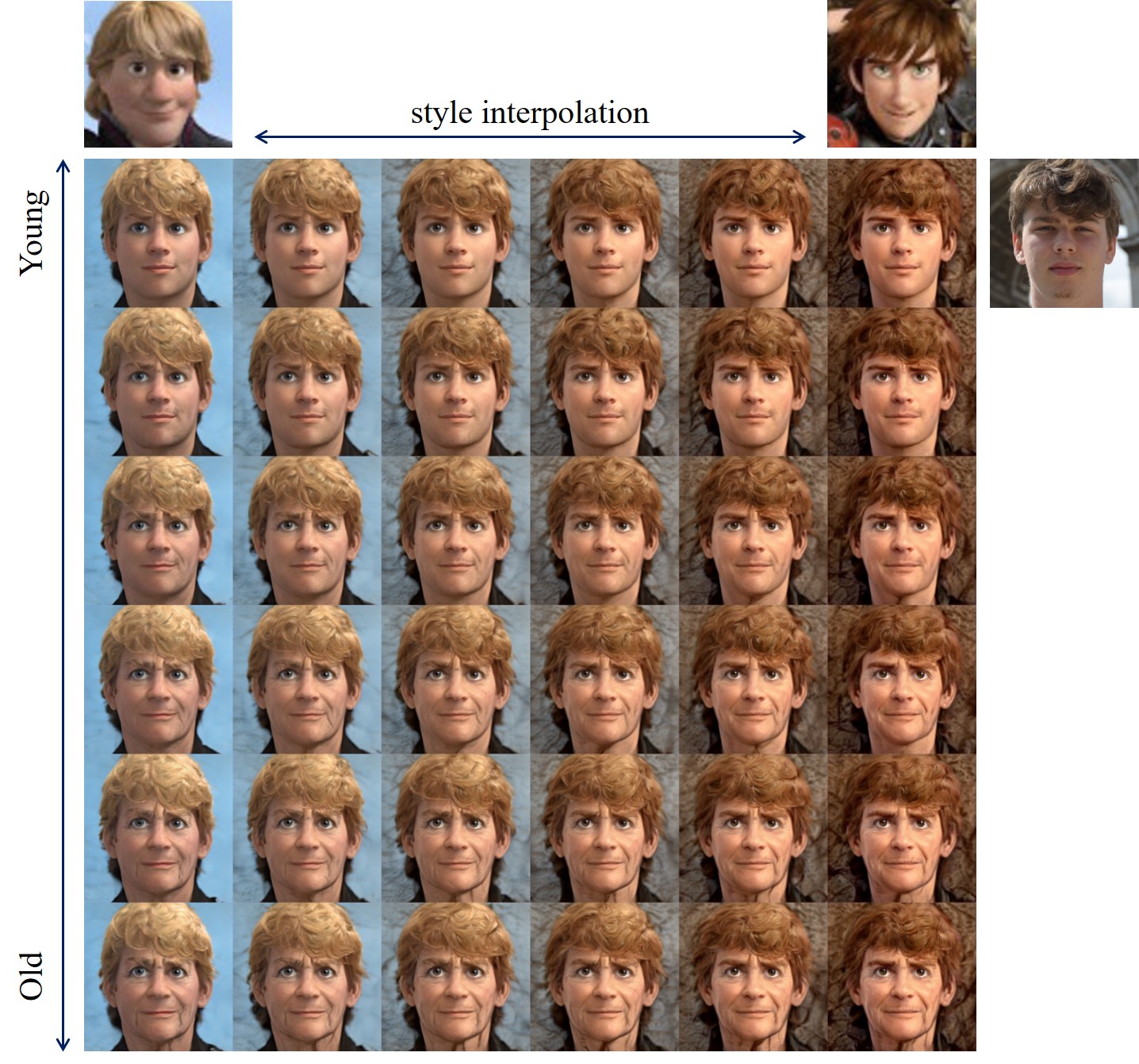}
  \caption{Style-Age interpolation. We interpolate two style images while varying the target age from the young to old direction.}
  \label{fig:style_age_interpolation}
\end{figure}

StyleGAN is known for its ability to interpolate latent vectors. Similarly, \textit{ToonAging} inherits this capability, allowing it to perform interpolation between two style references. Additionally, it can simultaneously transform the target age, creating more satisfying images. As depicted in Fig. \ref{fig:style_age_interpolation}, our method can successfully conduct latent vector interpolation that encompasses both style and age-related attributes.
\subsection{Re-Aging with Reference Image}

\begin{figure}[!t]
  \centering
  
  \includegraphics[width=0.9\linewidth]{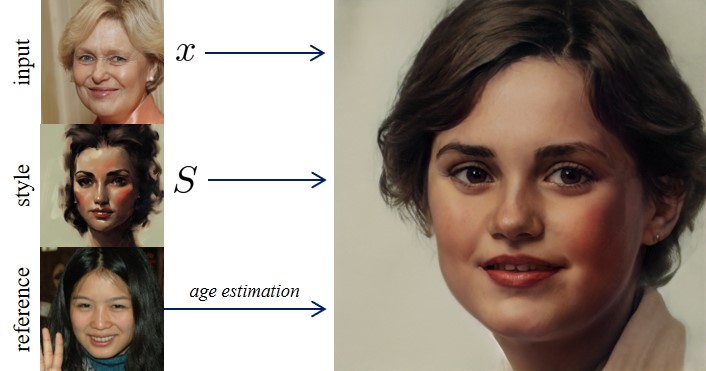}
  \caption{Extension of \text{ToonAging} for 'Age Reference' with an age estimator. Given an input PR image $x$ and style $S$, the reference image is predicted by an age classifier to estimate the target age.}
  \label{fig:extension}
\end{figure}

Conventionally, re-aging involves using a target age as input. We propose an alternative by introducing an 'Age Reference,' a reference image for the target age used in conjunction with the reference style. We estimate the target age through the age reference using the off-the-shelf age classifier DEX \cite{rothe2015dex}. Given that the perception of age can be subjective, an age reference can provide a more visually realistic target for re-aging, leading to more natural results compared to using a single number (target age). We have presented an example of age reference in Fig. \ref{fig:extension}.

\begin{figure}[t]
 
  \includegraphics[width=1\linewidth]{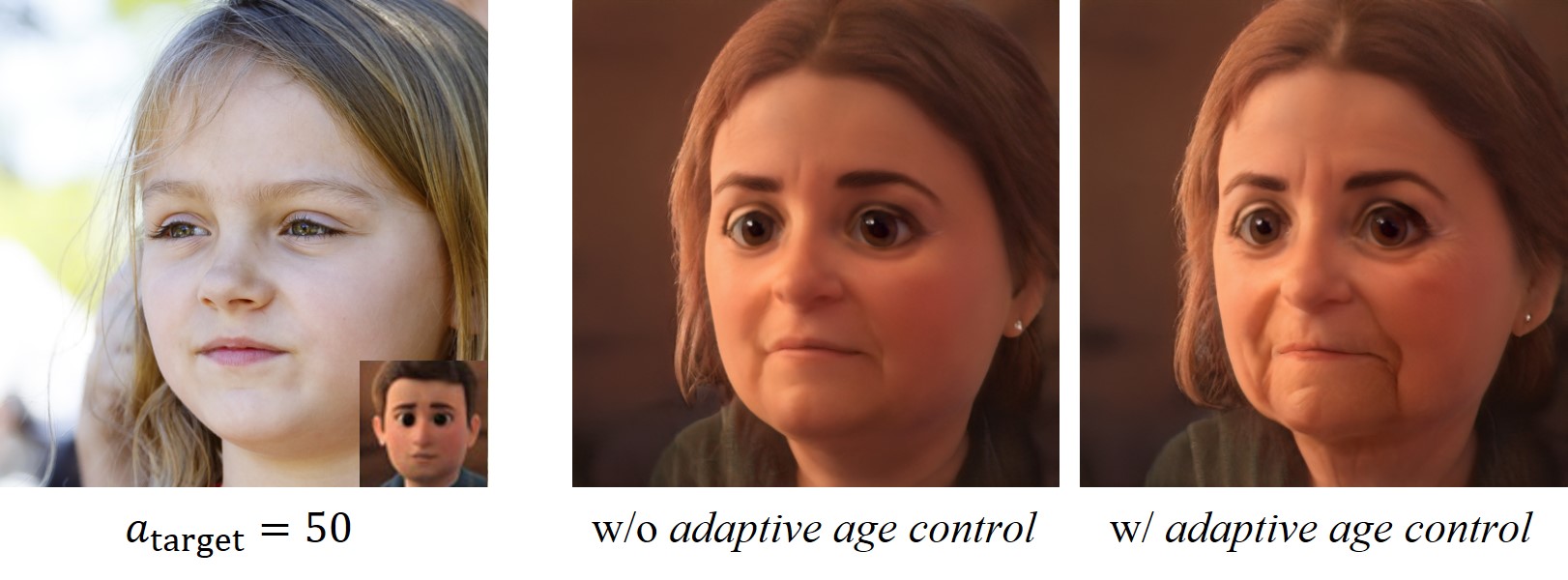}
  \caption{Advantage of \textit{Adaptive Age Control}: With \textit{adaptive age control}, users can effectively manipulate the aging effects.}
  
  \label{fig:our_adaptive}
\end{figure}

\subsection{Adaptive Age Control}

Although our method produces plausible results, we extend it to give more user control. We gain insights from section \ref{sec:latentfusionanalysis} and introduce an \textit{adaptive age control} method that can be defined as:

\begin{equation}
\hat{a}_{\text{target}} = a_{\text{target}} \times \left ( \frac{\Sigma_{i=1}^m \textbf{w}_i}{m} \right ) ^{-1},
\end{equation}

\noindent where $\hat{a}_{\text{target}}$ denotes the new target age. This allows us to modify the apparent age of the resulting image without compromising the style, bringing it closer to the target age, as illustrated in Fig. \ref{fig:our_adaptive}. It is evident that adaptive control exhibits more realistic output with a reduced perceptual age gap between the apparent age and the target age.

\subsection{User Study Sample}
\begin{figure}[h]
 \centering
   \includegraphics[width=0.95\linewidth]{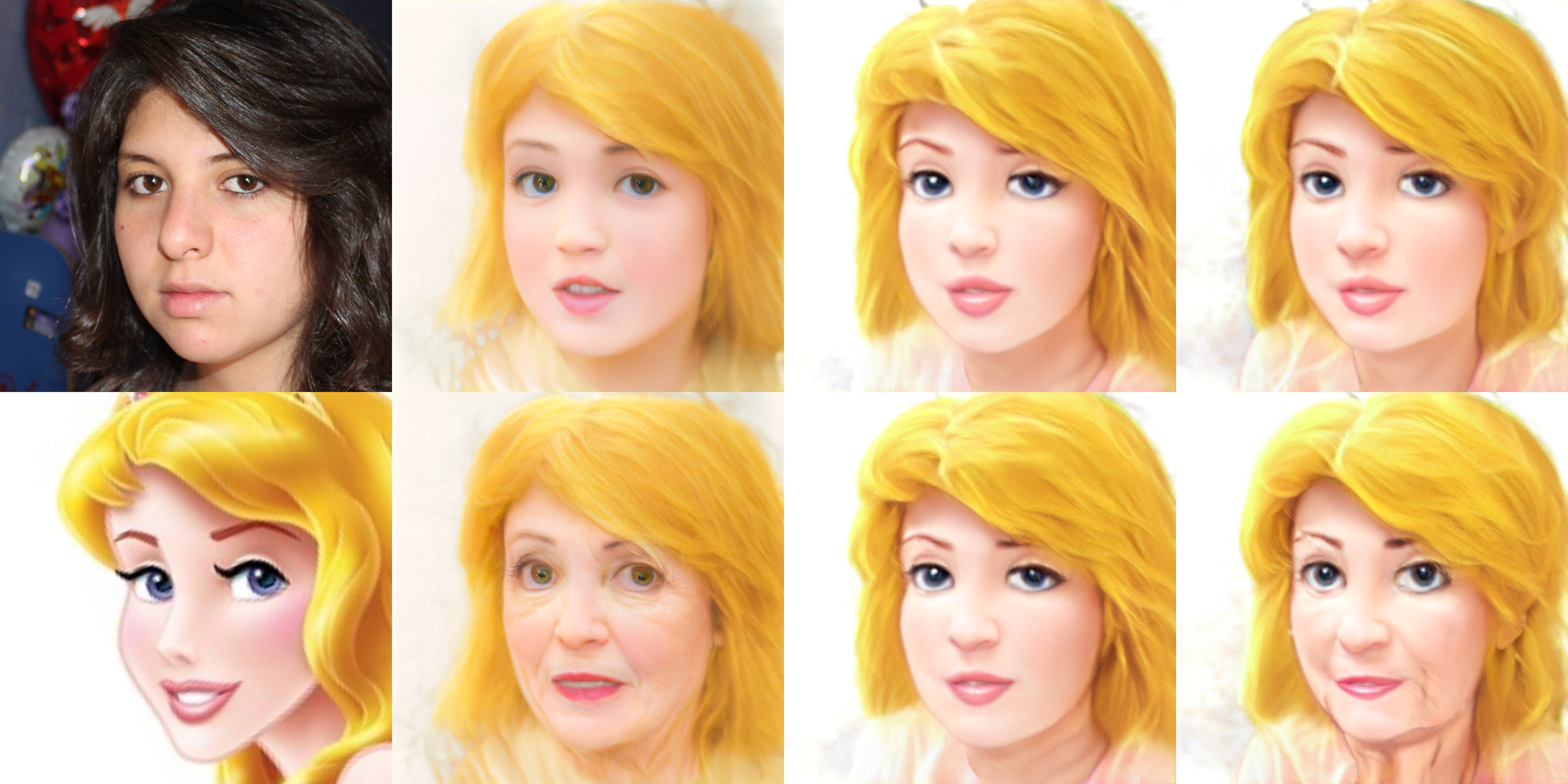}
  \caption{Sample image from the user study.}  
  \label{fig:user_study}
\end{figure}

Fig. \ref{fig:user_study} shows a sample image of the user study. We randomly shuffled the sequence to avoid any bias.

\subsection{Video ToonAging}
\begin{figure*}[!t]
  \centering
  \includegraphics[height=0.9\textheight,width=0.9\textwidth]{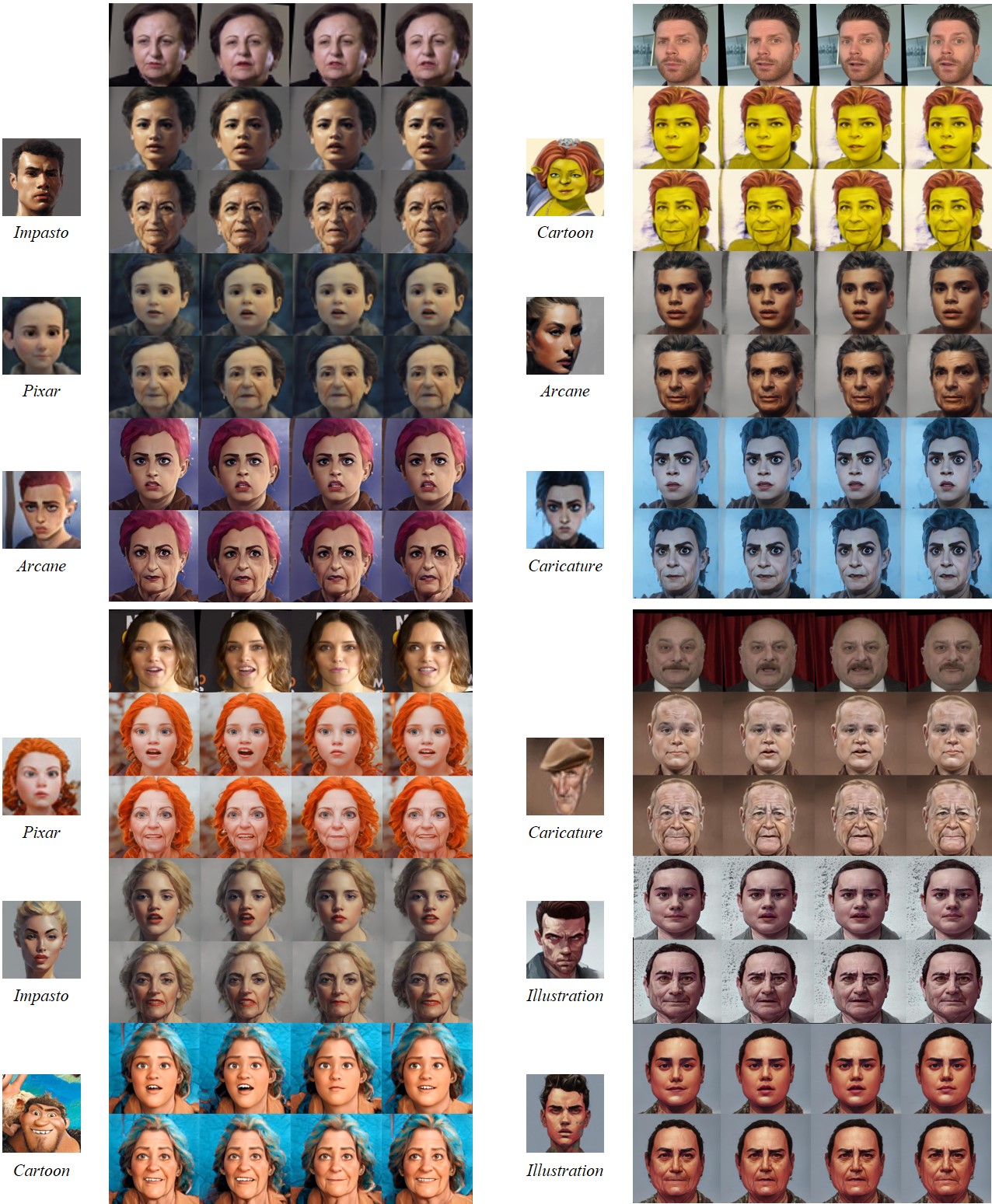}
  \caption{Video \textit{ToonAging}. In each set of results, the upper row denotes the young direction, and the lower row denotes the old direction.}
  \label{fig:vfhq_video_samples}
\end{figure*}

To evaluate the practicality of our \textit{ToonAging} method, we extended its application to video content. For this purpose, the VFHQ dataset \cite{xie2022vfhq} was chosen due to its high-quality and diverse video samples. The results, depicted in Fig. \ref{fig:vfhq_video_samples}, demonstrate that our approach effectively maintains the original facial expressions while concurrently modifying the style and perceived age to align with the designated reference style and target age. This experiment underscores the robustness of the \textit{ToonAging} technique in preserving essential facial dynamics and expressions across different media formats, showcasing its adaptability and effectiveness in video-based age transformation and style transfer.

\subsection{Ethical Statement}

The primary motivation of this work is for the entertainment industry, such as animated movies or personal usage. We highly discourage users for potential misuse of our method.

\subsection{Limitations and Failure Cases}
\begin{figure}[t]
  \centering  
  \includegraphics[width=1\linewidth]{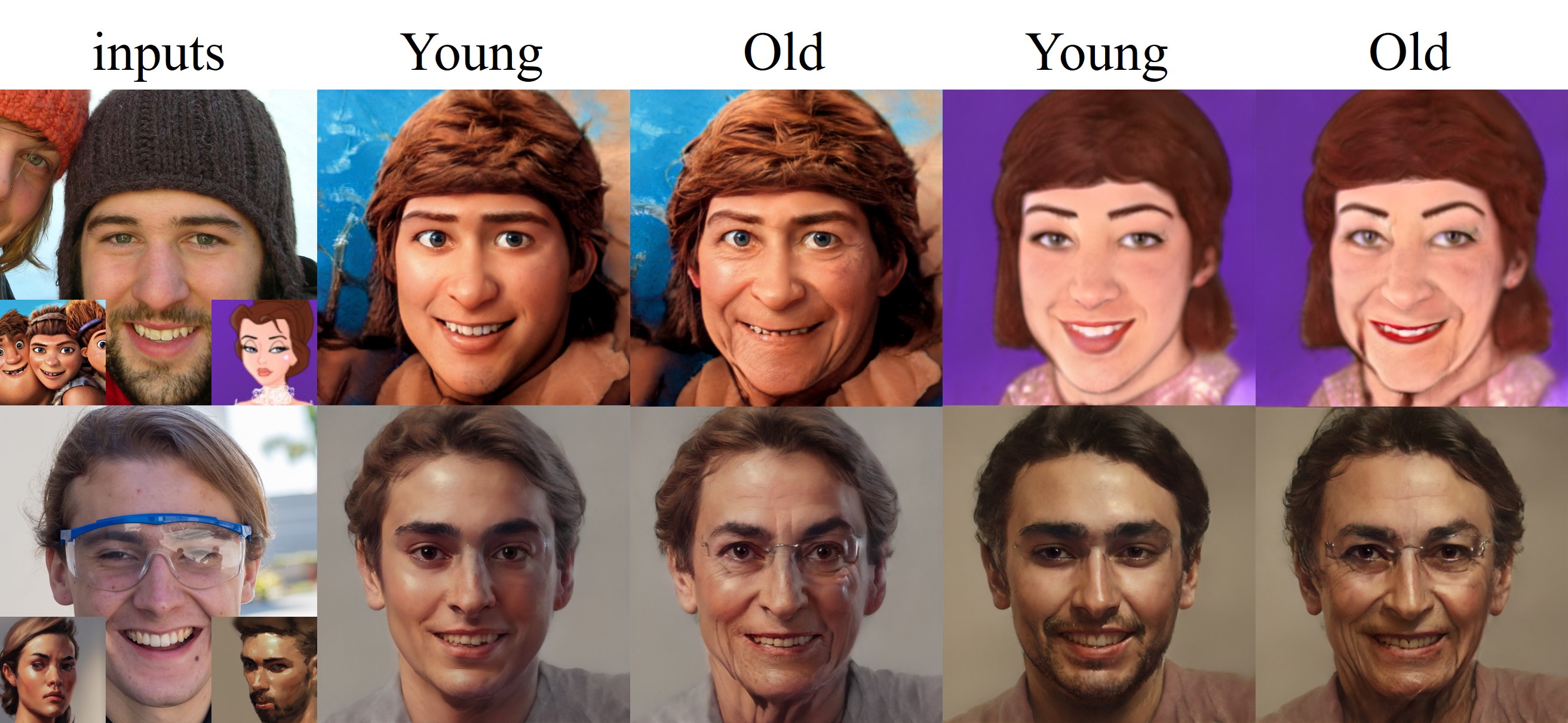}
  \caption{Failure cases of our method.}
  \label{fig:our_limitations}
\end{figure}

Despite generating high-quality results, our method fails to preserve the structure of the input image in some cases. These failure instances are presented in Fig. \ref{fig:our_limitations}. In the first row, we observe that the beanie dominates both style images, affecting both resultant outputs. Similarly, in the second row, glasses dominate the stylization process, leading to artifacts in the resultant images.

\newpage

\end{document}